\documentclass[journal]{IEEEtran}

\ifCLASSINFOpdf

\else

\fi
\usepackage{natbib}
\usepackage{amsmath}
\usepackage{bm}
\usepackage{mathtools}
\DeclareMathOperator*{\argmax}{\arg\!\max}

\usepackage{bm}
\usepackage{amssymb}
\usepackage{graphicx}
\usepackage{graphics}
\usepackage{subfigure}
\usepackage{xcolor}
\definecolor{darkblue}{rgb}{0,0,0.5}
\usepackage{transparent}
\usepackage{multirow}
\usepackage{booktabs}
\usepackage{soul}
\usepackage{nth}
\newcommand{\bx}{\mathbf{x}}
\newcommand{\bX}{\mathbf{X}}
\newcommand{\by}{\mathbf{y}}
\newcommand{\bY}{\mathbf{Y}}
\newcommand{\ba}{\mathbf{a}}
\newcommand{\bs}{\mathbf{s}}

\begin{document}

\title{A Survey of Deep Network Solutions\\ for Learning Control in Robotics: \\From Reinforcement to Imitation}
% \title{Deep Learning in Mobile Robotics - from Perception to Control Systems: A Survey on Why and Why not}

\author{
Lei~Tai$^{*1}$,
Jingwei~Zhang$^{*2}$,
Ming~Liu$^1$,
Joschka~Boedecker$^2$,
Wolfram~Burgard$^2$,
\thanks{*indicates equal contribution.}
\thanks{$^1$Lei Tai and Ming Liu are with The Hong Kong University of Science and Technology.
        \tt\small \{ltai, eelium\}ust.hk}
\thanks{$^2$Jingwei Zhang, Joschka Boedecker and Wolfram Burgard are with University of Freiburg.
        \tt\small \{zhang, jboedeck, burgard\}@informatik.uni-freiburg.de}
}
%\thanks{This work was sponsored by the Research Grant Council of Hong Kong SAR Government,
%China, under project No. 16212815, 21202816 and National Natural Science Foundation of China No. 6140021318, Shenzhen Science, Technology and Innovation Commission (SZSTI) JCYJ20160428154842603 and JCYJ20160401100022706.}
%    }

\markboth{Journal of \LaTeX\ Class Files,~Vol.~14, No.~8, August~2015}%
{Shell \MakeLowercase{\textit{et al.}}: Bare Demo of IEEEtran.cls for IEEE Journals}

\maketitle

\begin{abstract}
Deep learning techniques have been widely applied,
achieving state-of-the-art results in various fields of study.
This survey focuses on deep learning solutions that target learning control policies for robotics applications.
We carry out our discussions on the two main paradigms for learning control with deep networks: \textit{Deep Reinforcement Learning} and \textit{Imitation Learning}.
For \textit{Deep Reinforcement Learning} (DRL),
we begin from traditional reinforcement learning algorithms,
showing how they are extended to the deep context and effective mechanisms that could be added on top of the DRL algorithms.
We then introduce representative works that utilize DRL to solve navigation and manipulation tasks in robotics.
We continue our discussion on methods addressing the challenge of the \textit{reality gap} for transferring DRL policies trained in simulation to real-world scenarios,
and summarize robotics simulation platforms for conducting DRL research.
For \textit{Imitation Leaning},
we go through its three main categories,
behavior cloning, inverse reinforcement learning and generative adversarial imitation learning, by introducing their formulations and their corresponding robotics applications.
Finally, we discuss the open challenges and research frontiers.
\end{abstract}


%Several parts to add

%-- %1. voting, vote to vote (oxford) and 3D convolution (cmu shichao yang)
%2. Deep reinforcement learning from peter abteel and sergy levine
%3. Deep inverse reinfocement learning
%  Deep autoencoder
%-- % 4. Place recognition from RSS by object detection
%-- % 5. Place recognition from Wolfram work adided by sementic segmentation
%	road detection

%6. ETH Mark's work
%7. Fangyi work
%-- %8. Liu yong's work
%9. Google deep mind learning to navigate
% --%10. Ian Lenz work in deep learning robotics
% --%11. Libo's work

%openning in perception: 3D convolution, voting convolution
%openning in control: deep inverse reinforcement learning, deep reinforcement learning

%%%%%%%%% BODY TEXT
\section{Introduction}
\label{sec:introduction}

\subsection{Deep Learning}
\label{sec:dl}

Deep learning,
as a solution for artificial intelligence that is capable of building progressively more abstract representations of input data,
plays an essential role in various fields of study \citep{Goodfellow-et-al-2016}.

From image classification \citep{krizhevsky2012imagenet,he2016deep,huang2017densely},
to semantic segmentation \citep{long2015fully,chen2016deeplab},
from playing Atari games at the human-level with only pixel inputs \citep{mnih2015human,mnih2016asynchronous},
to learning policies capable of driving real robotic systems in navigation \citep{zhu2017target,zhang2017deep,tai2017virtual} and manipulation \citep{levine2016end,yu2018one} tasks,
the learning power of deep networks drives the state-of-the-art in various research directions \citep{schmidhuber2015deep}.% of machine learning.

Recent years have witnessed a rapidly growing trend of utilizing deep learning techniques for robotics tasks.
Replacing hand-crafted features with learned hierarchical distributed deep features,
learning control policies directly from high-dimensional sensory inputs, the robotics community is making solid progress towards building fully autonomous intelligent systems.

\subsection{Deep Learning for Robotics: From Perception to Control}
\label{sec:dl-rob}
Autonomous intelligent robotics systems require two essential building blocks: perception and control.

The perception pipeline can be viewed as a passive procedure:
intelligent agents receive observations from the environment,
then infer desired properties or detect target quantities from those sensory inputs.
We refer readers to \citet{deng2014tutorial} and \citet{guo2016deep} for a comprehensive overview of deep learning techniques for perception.
Compared with pure perception, the problem of control for autonomous agents goes one step further,
seeking to actively interact with or influence the environment by conducting sequences of actions.
This active nature leads to the following major distinctions between perception and control, in terms of deep learning based approaches:

\textbf{Data distribution:} When learning perception through supervised learning techniques,
the training datasets are collected and labeled before the learning phase begins.
In this case, the data points can be viewed as being independently and identically distributed (i.i.d),
such that a direct mapping from the input to the labels can be learned via standard stochastic gradient descent methods and variants.
In contrast, for control, the datasets are collected in an online manner, which makes the data points sequential in nature: the consecutive observations received by the agent are temporally correlated since the agent actively influences the data distribution by the actions it takes. Ignoring this underlying temporal correlation would lead to compounding errors \citep{bagnell2015invitation}.

\textbf{Supervision signal:} The supervision for learning perception is often direct and strong,
in that each training sample is provided along with its ground truth label.
In control tasks, on the other hand, either only sparse reward signals are available when learning behaviors through \textit{deep reinforcement learning},
or the feedback is often delayed and not instantaneous,
even when demonstrations from experts are provided in the scenario of \textit{imitation learning},
since the credit for achieving a certain goal needs to be correctly assigned to all the actions taken along the trajectory.

\textbf{Data collection:} As discussed before, the dataset for perception can be collected off-line, while the dataset for control has to be collected in an on-line manner, since actions are actively involved in the learning process. This greatly limits the number of samples one can collect, since executing actions in the real world with real robotics systems is a relatively expensive procedure. In cases where the control policies are trained in simulation, the problem of the \textit{reality gap} arises when they are deployed in real-world scenarios, where the discrepancies between the modalities of the synthetic renderings and the real sensory readings impose major challenges.

Recognizing those distinctions, various deep learning based algorithms have been proposed to solve control for robotics.
In this survey, we review the deep learning approaches for control tasks based on their underlying learning paradigms,
and we carry out our discussion through the following sections:

\begin{itemize}
  \item  \textbf{Sec. \ref{sec:drlaaa}} Deep Reinforcement Learning
    \begin{itemize}
      \item \textbf{Sec. \ref{sec:rl-overview}} RL Overview
      \item \textbf{Sec. \ref{sec:rl-algorithms}} RL Algorithms
      \item \textbf{Sec. \ref{sec:drl-algorithms}} DRL Algorithms
      \item \textbf{Sec. \ref{sec:drl-mechanisms}} DRL Mechanisms
      \item \textbf{Sec. \ref{sec:drl-navigation}} DRL for Navigation
      \item \textbf{Sec. \ref{sec:drl-manipulation}} DRL for Manipulation
      \item \textbf{Sec. \ref{sec:reality-gap}} The Reality Gap: From Simulation to the Real World
      \item \textbf{Sec. \ref{sec:sim}} Simulation Platforms
    \end{itemize}
  \item \textbf{Sec. \ref{sec:imitaaa}} Imitation Learning
    \begin{itemize}
      \item \textbf{Sec. \ref{sec:imi-bc}} Behavior Cloning
      \item \textbf{Sec. \ref{sec:imi-irl}} Inverse Reinforcement Learning
      \item \textbf{Sec. \ref{sec:imi-gail}} Generative Adversarial Imitation Learning
    \end{itemize}
\end{itemize}

\section{Deep reinforcement learning}
\label{sec:drlaaa}
Being the first to stabilize large-scale reinforcement learning with deep convolutional neural networks as function approximators,
deep Q-networks (DQN) \citep{mnih2015human} have brought increased research and applications of deep reinforcement learning (DRL) methods.
In the following we first review the basic concepts and algorithms in traditional reinforcement learning (RL). Then we continue to the several most influential DRL algorithms and mechanisms, on the basis of which we discuss DRL solutions for robotics control, with an emphasis on navigation and manipulation applications.

\subsection{\textbf{RL Overview}}
\label{sec:rl-overview}
We formalize a robotics task (e.g., navigation, manipulation) as a \textit{Markov Decision Process} (MDP), in which the agent interacts with the environment through a sequence of observations, actions, and reward signals. An MDP is a $5-$tuple $\left< \mathcal{S}, \mathcal{A}, P, \mathcal{R}, \gamma \right>$:
\begin{itemize}
  \item $\mathcal{S}$: set of all states.
  \item $\mathcal{A}$: set of all actions.
  %\item $\mathcal{P}(\bs_{t+1}|\bs_{t}, \ba_{t})$: defines the probability of transisting to state $\bs_{t+1}$ from a state-action pair $(\bs_{t}, \ba_{t})$.
  %\item $\mathcal{P}(S_{t+1}|S_{t}, A_{t})$: defines the probability of transisting to a next state from a state-action pair.
  \item $\mathcal{P}$: the transition dynamics,
  where $P(\bs'|\bs, \ba)$ defines the distribution of the next state $\bs'$ by taking action $\ba$ in state $\bs$,
  where $\bs,\bs'\in\mathcal{S}, \ba\in\mathcal{A}$. We also denote the initial state distribution $P(\bs_0)$ as $\rho_0$.
  %\item $\mathcal{R}(\bs_{t}, \ba_{t})$: defines the instantaneous scalar reward received by the agent from a state-action pair $(\bs_{t}, \ba_{t})$.
  %\item $\mathcal{R}(S_{t}, A_{t})$: defines the instantaneous scalar reward received by the agent from a state-action pair.
  % \item $\mathcal{R}(\bs_t,\ba_t,\bs_{t+1})$ \hl{check again here}: defines the instantaneous scalar reward received by the agent from a state-action pair. Usually one is interested in the expected reward achieved with a state-action pair: $R_{t+1}(\bs_t,\ba_t)=\mathbf{E}_{\bs_{t+1}\sim\mathcal{P}(\cdot|\bs_t,\ba_t)}\left[\mathcal{R}(\bs_t,\ba_t,\bs_{t+1})\right]$; in the following descriptions, we mostly use $R_{t+1}$ to be short for $R_{t+1}(\bs_t,\ba_t)$.
  \item $\mathcal{R}$: set of all possible rewards.
  In the following, we denote the instantaneous scalar reward received by the agent by taking action $\ba_t$ from state $\bs_t$ as $R_{t+1}(\bs_t,\ba_t)$,
  and use $R_{t+1}$ as short for $R_{t+1}(\bs_t,\ba_t)$.
  There also exist other definitions of the reward function that depend only on the state itself,
  in which $R(\bs)$ refers to the reward signal that the agent receives by arriving at state $\bs$.
  In some of the following discussions, the negative counterpart of the reward function,
  the cost function, is used, and is denoted as $c(\bs)$.
  \item $\gamma$: a discount factor in the range of $[0,1]$.
\end{itemize}

%\hl{TODO: make two figures out of this. fig:loop: showing the loop with gazebo,...in the environment, turtlrbot,quadrotor,manipulator in the agent box; fig:drl-nets: show the two approximator types}
\begin{figure}[h]
    \centering
    \includegraphics[width=0.9\columnwidth]{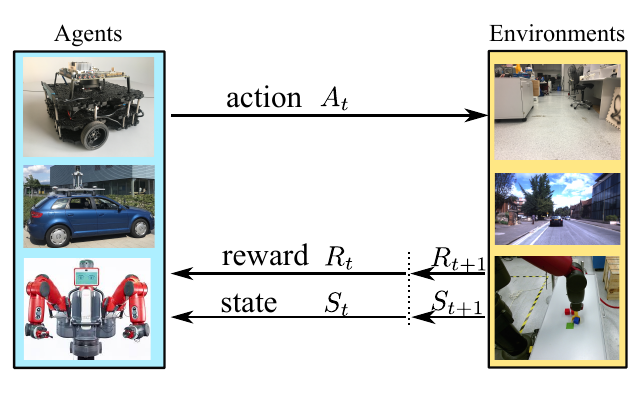}
    \caption{The reinforcement learning loop in the context of robotics.
    In state $\bs_{t}$, the autonomous agent takes and action $\ba_{t}$,
    receives a reward $R_{t+1}$, and transits to the next state $\bs_{t+1}$.}
    %\caption{A brief illustration of deep reinforcement learning methods. The upper part shows the standard reinforcement learning loop in which the agent interacts with the environment. The lower part shows the two different formulations of reinforcement learning algorithms, the deep networks are used as function approximators for the (action-)value function in value-based methods, or for representing the policy in policy-optimization methods.}
    \label{fig:loop}
\end{figure}

In an MDP, the agent takes an action $\ba_{t}$ in state $\bs_{t}$, receives a reward $R_{t+1}$, and transits to the next state $\bs_{t+1}$ following the transition dynamics $\mathcal{P}(\bs_{t+1}|\bs_{t},\ba_{t})$. This process in the context of robotics is depicted in Fig.\ref{fig:loop}.

In robotics, we mainly consider \textit{episodic} MDPs, where there exists a \textit{terminal state} (e.g., a mobile ground vehicle reaches a certain goal location, a manipulator successfully grabs a red cup) that, once reached, terminates the current \textit{episode}. Also for an \textit{episodic} MDP with a time horizon of $T$, an \textit{episode} will still be terminated after a maximum of $T$ time steps, even if by then the \textit{terminal state} has not yet been reached.

Another point worth mentioning is the \textit{partial observability}.
In a robotics task, an autonomous agent perceives the world with its onboard sensor (e.g., RGB/depth camera, IMU, laser range sensor, 3D Lidar),
receiving one observation
%$\mathbf{x}_{t}$ (e.g., one color/depth image, one laser scan, one ... point cloud)
per time step. However, simply representing $\bs_{t}$ by $\mathbf{x}_{t}$ often does not satisfy the \textit{Markov} property: one such sensor reading can hardly capture all the necessary information for the agent to make decisions in the future,
in which case the underlying procedure is called a Partial Observeble MDP (POMDP).
This is often dealt with by either stacking several (e.g, $N$) consecutive observations $\{ \mathbf{x}_{t-N+1}, \mathbf{x}_{t-N+2}, \dots, \mathbf{x}_{t} \}$ to represent $\bs_{t}$, or by feeding $\mathbf{x}_{t}$ into a \textit{recurrent} neural network instead of a \textit{feed forward} one,
such that the past information is naturally accounted with (e.g., by the cell state when using the long short-term memories (LSTMs).

Reinforcement learning agents are designed to learn from interactions how to behave to achieve a certain goal \citep{sutton1998reinforcement}.
More precisely, here the objective of learning is to maximize the \textit{expected discounted return},
where the \textit{discounted return} is defined as follows:
\begin{align}
  G_{t}
&=
  R_{t+1} + \gamma R_{t+2} + \gamma^2 R_{t+3} + \dots + \gamma^{T-t-1} R_{T}
\\&=
  \sum_{k=t}^{T} \gamma^{k-t} R_{k+1}.
\end{align}

To solve control, two important definitions are introduced:
\begin{itemize}
  \item \textbf{Policies}: $\pi, \mu$
    \begin{itemize}
      \item $\pi(\ba|\bs)$: stochastic policy, where actions are drawn from a probability distribution defined by $\pi(\ba|\bs)$.
      \item $\mu(\bs)$: determinstic policy, where actions are deterministically selected for a given state $\bs$.
    \end{itemize}
  \item \textbf{Value functions}: $V, Q$
    \begin{itemize}
      \item $V^{\pi}(\bs)$: state-value function, defined as the expected return when starting from state $\bs$ and following policy $\pi$ thereafter: % estimates how good it is to be in a particular state $\bs$ following policy $\pi$
      \begin{align}
        V^{\pi}(\bs)
      &=
        \mathbb{E}_{\pi}\left[ G_t | \bs_t=\bs \right]
      \\&=
        \mathbb{E}_{\pi}\left[ \sum_{k=t}^{T}\gamma^{k-t} R_{k+1} | \bs_t=\bs \right].
        \label{equ:value-reward}
      \end{align}
      \item $Q^{\pi}(\bs, \ba)$: action-value function,
      defined as the expected return by taking the action $\ba$ from state $\bs$,
      then following $\pi$ thereafter: % estimates how good it is to take a particular action $\ba$ in a particular state $\bs$ following policy $\pi$
      \begin{align}
        Q^{\pi}(\bs,\ba)
      &=
        \mathbb{E}_{\pi}\left[ G_t | \bs_t=\bs, \ba_t=\ba \right]
      \\&=
        \mathbb{E}_{\pi}\left[ \sum_{k=t}^{T}\gamma^{k-t} R_{k+1} | \bs_t=\bs, \ba_t=\ba \right].
        \label{equ:q-reward}
      \end{align}
      \item $Q^{*}(\bs, \ba)$: optimal value function (We omit the case for state-value function $V$ here since the action-value function $Q$ is a much more effective representation for control.):
      \begin{align}
%        V_{*}(\bs)
%      &=
%        \max_{\pi} V_{\pi}(\bs)
%      \\
        Q^{*}(\bs,\ba)
      &=
        \max_{\pi} Q^{\pi}(\bs,\ba).
      \end{align}
      \item $\pi^{*}(\ba|\bs)$: optimal policy:
      \begin{align}
        \pi^{*}(\ba|\bs)
      &=
        \argmax_{\ba} Q^{*}(\bs,\ba).
      \end{align}
    \end{itemize}
\end{itemize}

\subsection{\textbf{RL Algorithms}}
\label{sec:rl-algorithms}

With the definitions of the core components,
we now continue to discuss the different classes of RL algorithms.
We emphasize those methods that have been extended with deep learning variants.

% \begin{itemize}
%   \item \textbf{Value-based Methods}:
\subsubsection{\textbf{Value-based Methods}}
These methods are based on estimating the values of being in a given state,
then extracting the control policies from the estimated values.
The recursive value estimation procedures are based on the \textit{Bellman Equations}.
Below, we list the \textit{Bellman Expectation Equation} (Eq. \ref{equ:expectation}) and the \textit{Bellman Optimality Equation} (Eq. \ref{equ:optimality}):
\begin{align}
  Q^{\pi}(\bs,\ba)
  &=
  \mathbb{E}_{\pi}\left[ R_{t+1} + \gamma Q^{\pi}(\bs_{t+1},\ba_{t+1}) | \bs_t=\bs,\ba_t=\ba \right],
  \label{equ:expectation}
\\
  Q^{*}(\bs,\ba)
  &=
  \mathbb{E}\left[ R_{t+1} + \gamma \max_{\ba'} Q^{*}(\bs_{t+1},\ba') | \bs_t=\bs,\ba_t=\ba \right].
  \label{equ:optimality}
\end{align}

Following the formulations of Eq. \ref{equ:expectation} and Eq. \ref{equ:optimality} respectively,
we have the two most well-known value-based RL methods: \textit{SARSA} and \textit{Q-learning},
which follow the same recursive backup procedures, given as follows:
\begin{align}
  Q^{\pi}(\bs_{t},\ba_{t})
  &\leftarrow
  Q^{\pi}(\bs_{t},\ba_{t})+\alpha\delta_t,
  \label{equ:td}
\\
  \delta_t
  &=
  \by_t - Q^{\pi}(\bs_{t},\ba_{t}).
  \label{equ:tderror}
\end{align}

In this estimation procedure,
$Q$-values are recursively updated by a step size of $\alpha$ towards a target value $\by_t$.
$\delta_t$ is termed the \textit{td-error} (temporal difference error), and $\by_t$ the \textit{td-target}.

The difference between \textit{SARSA} and \textit{Q-learing} comes in their \textit{td-target's}. Below, we list the \textit{td-targets} for \textit{SARSA} and \textit{Q-learing} in Eq. \ref{equ:tdtargetS} and Eq. \ref{equ:tdtargetQ} respectively:
\begin{align}
  \by_t^{\textit{SARSA}} &= R_{t+1} + \gamma Q^{\pi}(\bs_{t+1},\ba_{t+1}),
  \label{equ:tdtargetS}
\\
  \by_t^{\textit{Q-learning}} &= R_{t+1} + \gamma \max_{\ba'} Q^{\pi}(\bs_{t+1},\ba').
  \label{equ:tdtargetQ}
\end{align}

\textit{SARSA} updates its $Q$-value estimates using the transitions generated by following the behavioural policy $\pi$, which makes \textit{SARSA} an \textit{on-policy} method; \textit{Q-learning}, on the other hand, is \textit{off-policy}, since its value estimations are updated not towards the behavioural policy, but towards a target optimal policy.

There are also other value-based methods,
such as \textit{Monte-Carlo control}, which uses the true return of complete trajectories as its update target instead of bootstrapping from old estimates,
and $\lambda$-variants, which mix the sample return and 1-step lookahead estimations.

A reformulation of the $Q$-value function,
the \textit{successor representation} \citep{dayan1993improving},
is also studied in the recent literature \citep{kulkarni2016deep, barreto2017successor, zhang2017deep}:
\begin{align}
  R_{t+1}(\bs_t,\ba_t)
&=
  \phi(\bs_t,\ba_t)^{\top}\cdot\omega,
  \label{equ:sr1}
\\
  Q^{\pi}(\bs,\ba)
&=
  \psi^{\pi}(\bs,\ba)^{\top}\cdot\omega,
  \label{equ:sr2}
\end{align}
where
\begin{align}
  \psi^{\pi}(\bs,\ba)
&=
  \mathbb{E}_{\pi}\left[ \sum_{k=t}^{T}\gamma^{k-t}\phi(\bs_k,\ba_k) | \bs_t=\bs,\ba_t=\ba \right],
  \label{equ:sr3}
\end{align}
is termed the \textit{successor feature}.
This line of formulation decouples the task specific reward estimation into the estimation of representative features $\phi(\cdot)$ and a reward weight $\omega$,
and the estimation of the expected occurrence of the features $\phi(\cdot)$ under specific world dynamics following a specific policy.
It combines the computational efficiency of model-free methods with the flexibility of some model-based methods.
We refer readers to \citet{dayan1993improving}, \citet{kulkarni2016deep}, \citet{barreto2017successor} and \citet{zhang2017deep} for more detailed discussions and extensions.

  % \item \textbf{Policy-based Methods}:
\subsubsection{\textbf{Policy-based Methods}}
Unlike value-based methods, policy-based methods do not maintain value estimations,
but work directly on policies. When it comes to high-dimensional or continuous action spaces, policy-based methods generally give much more effective solutions than value-based approaches. They can learn stochastic policies instead of just deterministic policies, and have better convergence properties.

Policy-based approaches operate on parameterized policies, and search for parameters that maximize the policy objective function. The policy search can be carried out in two paradigms: gradient-free \citep{fu2005simulation,szita2006learning} and gradient-based. We focus on the gradient descent methods from the gradient-based family as they remain the method of choice in recent studies. More formally, given policy $\pi_{\theta}(\cdot)$ with parameters $\theta$, policy optimization searches for the best $\theta$ that maximizes an objective function $\mathcal{J}(\pi_{\theta})$:
\begin{align}
  \mathcal{J}(\pi_{\theta})
&=
  \mathbb{E}_{\pi_{\theta}}[f_{\pi_{\theta}}(\cdot)].
  \label{equ:policy-obj}
\end{align}

Here, $f_{\pi_{\theta}}(\cdot)$ is a \textit{score function},
which judges the goodness of a policy.
There are multiple valid choices for the \textit{score function};
we refer readers to \cite{schulman2015high} for a full discussion.

The \textit{policy gradient} is defined as
\begin{align}
  \nabla_{\theta}\mathcal{J}(\pi_{\theta})
%&=
%  \mathbf{E}_{\pi_{\theta}}\left[ \nabla_{\theta}\log\pi_{\theta}(\cdot)Q^{\pi_{\theta}}(\bs,\ba) \right]
&=
  \mathbb{E}_{\pi_{\theta}}\left[ \nabla_{\theta}\log\pi_{\theta}\cdot f_{\pi_{\theta}}(\cdot) \right].
  \label{equ:pg}
\end{align}

Intuitively speaking, firstly, some actions,
experiences or trajectories are sampled following the current policy $\pi_{\theta}$ and the goodness of those samples is given by $f_{\pi_{\theta}}(\cdot)$,
the \textit{score function} and $\nabla_{\theta}\log\pi_{\theta}$ points out the direction in the parameter space that would lead to an increase of the probability of those actions being sampled.
Thus, by ascending along the \textit{policy gradient} given in Eq. \ref{equ:pg}, we end up with policies that are capable of generating samples with higher scores.

The standard REINFORCE algorithm \citep{williams1992simple}, a well-known method in RL, plugs in the sample return as the \textit{score function}:
\begin{align}
  f_{\pi_{\theta}}(\cdot)
&=
  G_t.
\end{align}

This algorithm, however, suffers from the very high variance. A common way to reduce the variance of the estimation while keeping it unbiased is by subtracting a \textit{baseline} $b(\bs)$ from the return:
\begin{align}
  f_{\pi_{\theta}}(\cdot)
&=
  G_t - b_t(\bs_t).
\label{equ:baseline}
\end{align}

A commonly used \textit{baseline} is a learned estimate of the \textit{state-value} function $V(\bs)$. This leads us to the \textit{actor-critic} class of algorithms, since it involves estimating the value functions along with policy search.

Before we go into \textit{actor-critic} methods, several details are worthy of pointing out.

Firstly, directly following the policy gradient might not be desirable in the robotics setting,
since hardware constraints and safety requirements should be carefully dealt with.
Popular approaches for cautious exploration include avoiding significant changes in the policy, or explicitly discouraging entering undesired regions in the state space \citep{deisenroth2013survey}.

We also note that, so far, we have only been discussing the policy gradient for the stochasitic polices,
which integrate over both the state and action spaces,
and might not be efficient in high-dimentional action spaces.
The deterministic policy gradient \citep{silver2014deterministic},
on the other hand, only requires integrating over the state space, which makes it a much more sample-efficient algorithm.
Below we list the \textit{stochastic policy gradient} (for $\pi_{\theta}(\ba|\bs)$,
Eq. \ref{equ:spg}) and the \textit{deterministic policy gradient} (for $\mu_{\theta}(\bs)$, Eq. \ref{equ:dpg})  when using the $Q$-value function as their \textit{score function}:

\begin{align}
  \nabla_{\theta}\mathcal{J}(\pi_{\theta})
&=
  \mathbb{E}_{\bs,\ba}\left[ \nabla_{\theta}\log\pi_{\theta}(\ba|\bs) \cdot Q^{\pi}(\bs,\ba) \right],
  \label{equ:spg}
\\
  \nabla_{\theta}\mathcal{J}(\mu_{\theta})
&=
  \mathbb{E}_{\bs}\left[ \nabla_{\theta}\mu_{\theta}(\bs) \cdot Q^{\mu}(\bs,\mu_{\theta}(\bs)) \right].
  \label{equ:dpg}
\end{align}

  % \item \textbf{Actor-critic Methods}:
\subsubsection{\textbf{Actor-critic Methods}}
Following on from the discussions of the \textit{policy-based} methods,
\textit{actor-critic} algorithms maintain an explicit representation of both the policy (the \textit{actor}) and the value estimates (the \textit{critic}). The most widely used \textit{actor-critic} algorithms use the following \textit{score function}:
\begin{align}
  f_{\pi_{\theta}}(\cdot)
&=
  Q^{\pi_{\theta}}(\bs_t,\ba_t) - V^{\pi_{\theta}}(\bs_t).
\label{equ:actor-critic}
\end{align}

Compared againest Eq. \ref{equ:baseline}, Eq. \ref{equ:actor-critic} replaces the return $G_t$ with its unbiased estimate $Q^{\pi_{\theta}}(\bs_t,\ba_t)$, and uses $V^{\pi_{\theta}}(\bs_t)$ as its \textit{baseline} function to reduce variance. In fact,
\begin{align}
  A(\bs,\ba)
&=
  Q(\bs,\ba) - V(\bs)
\label{equ:advantage}
\end{align}
is called the \textit{advantage} function, which estimates the advantage of taking a particular action $\ba$ in state $\bs$.

\subsubsection{\textbf{Integraing Planning and Learning}}
So far, we have been discussing \textit{model-free} methods where the agent is not provided with the underlying transition model and simply learns optimal behaviors from experiences.
There also exists another branch of \textit{model-based} algorithms where a model is learned from experiences.
with which the agent can interact and collect \textit{imaginary rollouts} \citep{sutton1991dyna},
It has also been extended with DRL methods \citep{weber2017imagination,pmlr-v78-kalweit17a}.
However, the need for learning a model brings in another source of approximation error,
and \textit{model-based} RL can only perform as well as the estimated model.
% \hl{related works???}
This problem might be partially dealt with by \textit{Model Predicted Control} (MPC) methods,
which are not a focus of this survey, so we will skip the details.
% \hl{cite???}\cite{mpc}.

% \end{itemize}

\subsection{\textbf{DRL Algorithms}}
\label{sec:drl-algorithms}

Recent successes of DRL have extended the aforementioned algorithms to the high-dimensional domain,
by deploying deep neural networks as powerful non-linear function approximators for the optimal value functions $V^*(\bs), Q^*(\bs,\ba), A^*(\bs,\ba)$,
and the optimal policies $\pi^*(\ba|\bs)$, $\mu^*(\bs)$.
They usually take the observations as input (e.g, raw pixel images from Atari emulators \citep{mnih2015human} or joint angles of robot arms \citep{}),
%, \hl{joint angles of ...}),
and output either the $Q$-values, from which greedy actions are selected, or policies that can be directly used to execute agents.
In the following, we cover the most influential DRL algorithms.

\subsubsection{\textbf{DQN} \citep{mnih2015human}}:
As a \textit{value-based} method, DQN approximates the optimal $Q$-value function with a deep convolutional neural network, called the deep \textit{Q-network}, whose weights we denote as $\theta^Q$: $Q(\bs,\ba;\theta^Q)\approx Q^*(\bs,\ba)$. In turn, the \textit{td-error} (Eq. \ref{equ:tderror}) and the \textit{td-target} (Eq. \ref{equ:tdtargetQ}) from the standard \textit{Q-learning} are adopted into:
\begin{align}
  \delta_t^{\text{DQN}}
&=
  \by_t^{\text{DQN}} - Q(\bs_{t},\ba_{t};\theta^Q_t),
  \label{equ:tderror-dqn}
\\
  \by_t^{\text{DQN}}
&=
  R_{t+1} + \gamma \max_{\ba'} Q(\bs_{t+1},\ba';\theta^{-}_t).
  \label{equ:tdtargetQ-dqn}
\end{align}
Then an update step is performed based on the following gradient calculation with a learning rate of $\alpha$:
\begin{align}
  \theta_{t+1}
&\leftarrow
  \theta_{t} - \alpha \cdot \left( \partial\left(\delta^{\text{DQN}}_{t}(\theta^{Q}_{t})\right)^2 / \partial{\theta^{Q}_{t}} \right).
\end{align}

Two main techniques have been proposed in DQN to stabilize learning: \textit{target-network} and \textit{experience replay}.

\textbf{\textit{Target-network}}: In Eq. \ref{equ:tdtargetQ-dqn}, the \textit{td-target} is computed using the output from a \textit{target-network} $\theta^{-}$, instead of the \textit{Q-network} $\theta^Q$.
The \textit{target-network} and the \textit{Q-network} share the same network architecture,
but only the weights of the \textit{Q-network} are learned and updated.
The weights of the \textit{Q-network} $\theta^Q$ are only periodically copied to the \textit{target-network} $\theta^{-}$. This reduces the correlations of the estimated $Q$-values with the target estimations.
There is also \textit{soft update} \citep{lillicrap2015continuous},
where a small portion of $\theta^Q$ are mixed into $\theta^{-}$ in every iteration,
instead of the \textit{hard update} used in the original DQN,
where $\theta^Q$ are directly and completely copied to $\theta^{-}$ every several (e.g, $10,000$) iterations.

\textbf{\textit{Experience replay}}: In this technique, instead of directly using the incoming frames from the online interactions,
the collected experiences are firstly stored into a \textit{replay memory}.
During training, random samples are drawn from the \textit{replay memory} ($4$ consecutive observations are stacked together to form a state, so as to deal with the \textit{partial observability}) to be fed into the network as mini-batches. This way, gradient descent methods from the supervised learning literature can be safely used, to minimize the min-squared error between the predicted $Q$-values (output by the \textit{Q-network}) and the target $Q$-values (output by the \textit{target-network}). \textit{Experience replay} thereby removes the temporal correlations in the consecutive observations, and smoothes over changes in the online data distribution.

Further techniques have been proposed on the basis of DQN to stabilize learning and improve efficiency: Double DQN \citep{van2016deep} and Dueling DQN \citep{wang2015dueling}.

For Double DQN \citep{van2016deep}, the greedy action is chosen based on the output from $\theta^Q$ (the original DQN uses $\theta^{-}$),
then the target $Q$-value of the chosen greedy action is computed using $\theta^{-}$ (Eq. \ref{equ:tdtargetQ-double}). This prevents overoptimistic value estimates and avoids upward bias:
\begin{align}
  \by_t^{\text{Double}}
&=
  R_{t+1} + \gamma Q(\bs_{t+1},\argmax_{\ba'}Q(\bs_{t+1},\ba';\theta^Q_t);\theta^{-}_t).
  \label{equ:tdtargetQ-double}
\end{align}

For Dueling DQN \citep{wang2015dueling}, two output heads are used to estimate the \textit{state-value} $V$ and the \textit{advantage} $A$ respectively for each action. This helps the agent to efficiently learn which states are valuable, without having to learn the effect of each action for each state.

\subsubsection{\textbf{DDPG} \citep{lillicrap2015continuous}}:
DQN can deal with high-dimensional state spaces, but is only capable of handling discrete and low-dimensional action spaces.
The deep deterministic policy gradient (DDPG) combines techniques from DQN with \textit{actor-critic} methods, targeting solving continuous control tasks from raw pixels inputs.

If we write out the expectation in Eq. \ref{equ:expectation} for the \textit{stochasitic policy} $\pi(\ba|\bs)$ and \textit{deterministic policy} $\mu(\bs)$, we get ($E$ represents the environment that the agent is interacting with) %\hl{move Q to the beginning of lines}
\begin{align}
    &Q^{\pi}(\bs_t,\ba_t)
=\nonumber\\
    &\mathbb{E}_{R_{t+1},\bs_{t+1}\sim E}\left[ R_{t+1} + \gamma \mathbb{E}_{\ba_{t+1}\sim{\pi}} \left[ Q^{\pi}(\bs_{t+1},\ba_{t+1})\right] \right],
    \label{equ:expectation-s}
\\
    &Q^{\mu}(\bs_t,\ba_t)
=\nonumber\\
    &\mathbb{E}_{R_{t+1},\bs_{t+1}\sim E}\left[ R_{t+1} + \gamma Q^{\mu}(\bs_{t+1},\mu(\bs_{t+1}))\right].
    \label{equ:expectation-d}
\end{align}

DDPG represents the $Q$-value estimates with $\theta^Q$,
and the deterministic policy with $\theta^{\mu}$.
$\theta^{\mu}$ is learned via the DPG given in Eq. \ref{equ:dpg}, and $\theta^Q$ is learned following Eq. \ref{equ:expectation-d}.
(Note that different from DQN, where the dependence of the $Q$-value on $\ba$ is represented by outputting one value for each action, the $Q$-network in DDPG deals with this dependence by taking the action as input for $\theta^Q$.)

\subsubsection{\textbf{NAF} \citep{gu2016continuous}}:
Normalized advantage function offers another way to enable $Q$-learning in continuous action spaces with deep neural networks and is considerably simpler than DDPG.
For continuous action problems, standard $Q$-learning is not easily directly applicable,
as it requires maximizing a complex non-linear function for determining the greedy action.
The key idea in NAF is to represent the $Q$-value function $Q(\bs,\ba)$ in such a way that its maximum $\argmax_{\ba}Q(\bs,\ba)$ can be easily analytically determined during the $Q$-learning update.

NAF uses the same techniques of \textit{target network} and \textit{experience replay} as DQN,
but differs in the network outputs. Instead of directly outputting the $Q$-value estimates,
its last hidden layer is connected to three output heads: $\theta^V$, $\theta^{\mu}$ and $\theta^{L}$.
$\theta^V$ represents the \textit{state value} $V(\bs)$,
while $\theta^{\mu}$ and $\theta^{L}$ are used for estimating the \textit{advantage} $A(\bs,\ba)$;
then $Q(\bs,\ba)$ can be computed according to Eq. \ref{equ:advantage}.
To give a specific example, e.g, if both $\theta^{\mu}$ and $\theta^L$ are represented with linear layers,
with the number of outputs of $\theta^L$ being the square of that of $\theta^{\mu}$ (equal to the action dimensions),
then the output of $\theta^L$ is first reshaped into a matrix, from which $L(\bs;\theta^L)$, being the lower-triangular of that matrix, is extracted, with the diagonal terms exponentiated. Then the \textit{advantage} can be estimated by
\begin{align}
    &A(\bs,\ba;\theta^{\mu},\theta^L)
\nonumber\\&=
    -\frac{1}{2}
    \left( \ba-\mu(\bs;\theta^{\mu}) \right)^T
    P(\bs;\theta^L)
    \left( \ba-\mu(\bs;\theta^{\mu}) \right),
\end{align}
where
\begin{align}
    P(\bs;\theta^L)
&=
    L(\bs;\theta^L) L(\bs;\theta^L)^T.
\end{align}

Although this representation is more restritive than a general network approximator, the greedy action for the $Q$-value is always directly given by $\mu(\bs;\theta^{\mu})$.
An asynchronous version of NAF has also been proposed \citep{gu2017deep}.

\subsubsection{\textbf{A3C} \citep{mnih2016asynchronous}}:
Minh et. al. proposed several asynchronous DRL algorithms. They deploy multiple actor-learners to collect experiences on multiple instances of the environment, while each actor-learner accumulates gradients calculated from its own collected samples w.r.t. its own set of network parameters $\theta$; these gradients are used to update the weights of a shared model $\bm{\theta}$.

The most effective one, A3C (asynchronous advantage actor-critic), which has been very influential and become a standard baseline in recent DRL research,
maintains a policy representation $\pi(\ba|\bs;\bm{\theta}^{\pi})$ and a value estimate $V(\bs;\bm{\theta}^V)$.
It uses the \textit{advantage} function as the \textit{score fucntion} in its policy gradient,
which is estimated using a mixture of $n$-step returns by each actor-learner.
To be more specific, each actor-learner thread spawns its own copy of the environment and collects rollouts of experiences up to $T_{\max}$ (e.g, $20$) steps.
After an actor-learner completes a segment of a rollout,
it accumulates gradients from the experience of every time step contained in the rollout $\left\{ 0, 1, \cdots, t, \cdots, T \right\}$ by first estimating the \textit{advantage} function (e.g., for time step $t$) according to the following formulation
\begin{align}
  &A(\bs_t,\ba_t;\theta^{\pi},\theta^{V})
=\nonumber\\
  & \left[ \sum_{k=t}^{T-1} \left[ \gamma^{k-t} R_{k+1} \right] +\gamma^{T-t} V(\bs_{T};\theta^V) - V(\bs_{t};\theta^V); \theta^{\pi} \right],
\end{align}
then calculating the corresponding gradients w.r.t. the its current set of network parameters $\theta^{\pi},\theta^V$, which are then used to update the shared model $\bm{\theta}^{\pi},\bm{\theta}^V$:
\begin{align}
  d\bm{\theta}^{\pi}
&\leftarrow
  d\bm{\theta}^{\pi}
+
  \nabla_{\theta^{\pi}}\log\pi(\ba_t|\bs_t;\theta^{\pi})A(\bs_t,\ba_t;\theta^{\pi},\theta^{V}),
\\
  d\bm{\theta}^{V}
&\leftarrow
  d\bm{\theta}^{V} + \partial A(\bs_t,\ba_t;\theta^{\pi},\theta^{V})^2/ \partial\theta^{V}.
\end{align}

The parallelization greatly stabilizes the update of the parameters as the samples collected by different actor-learners at the same time are much less correlated,
which eliminates the requirement for keeping a \textit{replay memory}.
Also by running different exploration policies in different threads,
the learners are very likely to explore different parts of the state space.
Due to it being highly efficient, lightweight and conceptually simple, A3C is considered as a standard starting point in recent DRL research.

\subsubsection{\textbf{A2C} \citep{wang2016learning,wu2017scalable}}:
Some recent works found that the asynchrony in A3C does not necessarily lead to improved performance compared to the synchronous version: A2C.
Different from A3C, A2C waits for each actor to finish its segment of experience before performing an update,
which is averaged over all actors. This detail allows for effective GPU implementation.

\subsubsection{\textbf{GPS} \citep{levine2013guided}}:
As a model-based policy search algorithm, Guided Policy Search (GPS) is relatively sample efficient. GPS starts from guiding samples generated from some initial optimal control policies and augmented from samples generated from the current policy, from which at every iteration a set of training trajectories are sampled to optimize the current policy with supervised learning. The updated policy is then added as an additional cost term to bound the change in the policy, with which the trajectory optimization is performed again (e.g., with an LQR solver).

\subsubsection{\textbf{TRPO} \citep{schulman2015trust}}:
By making several approximations to the theoretically justified scheme, \citet{schulman2015trust} proposed a practical algorithm for optimizing large nonlinear policies, with guaranteed monotonic improvement.

To illustrate the algorithm, let us first define the \textit{expected discounted cost} for an infinite horizon MDP, which replaces the reward function $R$ in the \textit{expected discounted return} with the cost function $c$:
\begin{align}
  % \eta^{\pi}(\bs_0)
  \eta(\pi)
&=
  \mathbb{E}_{\pi} \left[ \sum_{t=0}^{\infty} \gamma^{t}c(\bs_t) | \bs_0\sim\rho_0 \right].
\end{align}

In turn, we can rewrite the definitions for the state-value functions Eq. \ref{equ:value-reward} and action-value functions Eq. \ref{equ:q-reward} in terms of the cost function $c$:
\begin{align}
  V^{\pi}(\bs)
&=
  \mathbb{E}_{\pi}\left[ \sum_{k=t}^{\infty}\gamma^{k-t} c(\bs_{k}) | \bs_t=\bs \right],
  \label{equ:value-cost}
\\
  Q^{\pi}(\bs,\ba)
&=
  \mathbb{E}_{\pi}\left[ \sum_{k=t}^{\infty}\gamma^{k-t} c(\bs_{k}) | \bs_t=\bs, \ba_t=\ba \right],
  \label{equ:q-cost}
\end{align}
and in turn, we have the advantage function:
\begin{align}
  A^{\pi}(\bs,\ba)
&=
  Q^{\pi}(\bs,\ba)-V^{\pi}(\bs).
\end{align}
Since we are looking for a step size for the policy update that can guarantee a monotonic improvement from an old policy $\pi_{\text{old}}$ to an updated policy $\pi$, it is beneficial to write the expected cost of $\pi$ in terms of that of $\pi_{\text{old}}$, which leads to the following identity:
\begin{align}
  % \eta^{\pi}(\bs_0)
  \eta(\pi)
&=
  % \eta^{\pi_{\text{old}}}(\bs_0)
  \eta(\pi_{\text{old}})
+
  \mathbb{E}_{\pi} \left[ \sum_{t=0}^{\infty} \gamma^t A^{\pi_{\text{old}}}(\bs_t,\ba_t) | \bs_0\sim\rho_0 \right].
  \label{equ:cost-rewrite}
\end{align}

Before continueing, we denote the (unnormalized) discounted visitation frequencies for state $\bs$ under policy $\pi$ as $\rho^{\pi}(\bs)$, more formally,
\begin{align}
  \rho^{\pi}(\bs)
&=
  \left( P(\bs_0=\bs) + \gamma P(\bs_1=\bs) + \gamma^2 P(\bs_2=\bs) + \cdots \right)
\nonumber\\&=
  \sum_{t=0}^{\infty}\gamma^{t}P(\bs_t=\bs),
\end{align}
where $\bs_0\sim\rho_0$, and the actions are selected following $\pi$.

Now, instead of summing over timesteps, if we sum over states, Eq. \ref{equ:cost-rewrite} can be rewritten as
\begin{align}
  % \eta^{\pi}(\bs_0)
  \eta(\pi)
&=
  \eta(\pi_{\text{old}})
+
  \sum_{\bs\sim\rho^{\pi}}\rho^{\pi}(\bs)\sum_{\ba\sim\pi}\pi(\ba|\bs)A^{\pi_{\text{old}}}(\bs,\ba).
  \label{equ:cost-rerewrite}
\end{align}

This equation indicates that $\eta$ is guaranteed to decrease or stay constant if the expected advantage at every state has a non-positive value.
Since Eq. \ref{equ:cost-rerewrite} is difficult to directly optimize, due to the complex dependency of $\rho^{\pi}$ on $\pi$, a local approximation ignoring changes in the state visitation density induced by the changes in the policy, that matches $\eta$ to the first order is introduced (the term $\eta(\pi)$ is left out here as it does not affect the solution to the underlying optimization problem):
\begin{align}
  L_{\pi_{\text{old}}}{(\pi)}
=
  % &\eta(\pi_{\text{old}})+
  \sum_{\bs\sim\rho^{\pi_{\text{old}}}}\rho^{\pi_{\text{old}}}(\bs)
  \sum_{\ba\sim\pi}\pi(\ba|\bs)
  A^{\pi_{\text{old}}}(\bs,\ba).
  % \sum_{\bs}\rho^{\pi}(\bs)\sum_{\ba}\pi(\ba|\bs)A^{\pi_{\text{old}}}(\bs,\ba)
  \label{equ:cost-approx1}
\end{align}

Standard policy gradient methods ascend on the $\nth{1}$ order gradient, where an increase on
$L_{\theta_{\text{old}}}{(\theta)}$
% $L_{\pi_{\text{old}}}{(\pi)}$
does not guarantee an increase in
$\eta({\pi_{\theta}})$
% $\eta({\pi})$
with large step sizes, due to the approximations made above.

TRPO extends the policy improvement bound in the mixture policies setting given by \citet{kakade2002approximately} to general stochastic policies,
and shows that
\begin{align}
  \eta(\pi)
&\leq
  L_{\pi_{\text{old}}}(\pi)
+
  CD_{\text{KL}}^{\max}(\pi_{\text{old}},\pi),
\label{equ:soft}\\
\end{align}
where
\begin{align}
  C
&=
  \frac{2\epsilon\gamma}{(1-\gamma)^2},
\label{equ:constant}\\
  \epsilon
&=
  \max_{\bs}\left| \mathbb{E}_{\ba\sim\pi}\left[ A^{\pi_{\text{old}}}(\bs,\ba) \right] \right|.
\end{align}
This means that by performing the following optimization (here we denote $L_{\theta_{\text{old}}}(\theta)\coloneqq L_{\pi_{\theta_{\text{old}}}}(\pi_{\theta})$) with parameterized policies),
we are guaranteed to improve the true objective $\eta$:
\begin{align}
% \begin{equation*}
% \begin{aligned}
& \underset{\theta}{\text{minimize}}
\left[ L_{\theta_{\text{old}}}(\theta)+CD_{\text{KL}}^{\max}(\pi_{\theta_{\text{old}}},\pi_{\theta}) \right].
% & & {\mathbb{E}}_{\bs\sim\rho^{\pi_{\theta{\text{old}}}},\ba\sim\pi_{\theta{\text{old}}}}
% \left[ \frac{\pi_{\theta}(\ba|\bs)}{\pi_{\theta{\text{old}}}(\ba|\bs)} Q^{\pi_{\theta{\text{old}}}}(\bs,\ba) \right]\\
% & \text{subject to}
% & & {\mathbb{E}}_{\bs\sim\rho^{\pi_{\theta{\text{old}}}}}
% \left[
% D_{\text{KL}}(\pi_{\theta_{\text{old}}}(\cdot|\bs) || \pi_{\theta}(\cdot|\bs))\right]
% \leq\delta.
% & &
% \end{aligned}
% \end{equation*}
\label{equ:soft}
\end{align}

However, if the penalty coefficient $C$, as calculated in Eq. \ref{equ:constant}, is used in practice, the step sizes will be very small.

% Now, we overload the notations in Eq. \ref{equ:cost-approx1} with parameterized policies, and replace the sum over actions by an importance sampling estimator (here we only discuss the case for \textit{single path} sampling, where $\pi_{\theta_{\text{old}}}$ is used to generate trajectories):
% \begin{align}
%   L_{\theta_{\text{old}}}{(\theta)}
% &=
%   \eta(\theta_{\text{old}})
% +
%   \sum_{\bs\sim\rho^{\pi_{\theta_{\text{old}}}}}\sum_{\ba\sim\pi_{\theta{\text{old}}}} \left[ \frac{\pi_{\theta}(\ba|\bs)}{\pi_{\theta_{\text{old}}}(\ba|\bs)} A^{\pi_{\theta_{\text{old}}}}(\bs,\ba) \right].
%   \label{equ:cost-approx2}
% \end{align}

To deal with this, TRPO first replaces the sum over actions in Eq. \ref{equ:cost-approx1} by an importance sampling estimator (here, we only discuss the case for \textit{single path} sampling, where $\pi_{\text{old}}$ is used to generate trajectories) ($A^{\pi_{\theta_{\text{old}}}}$ is replaced by $Q^{\pi_{\theta_{\text{old}}}}$ which only changes the objective by a constant, and the $Q$-values are to be replaced by empirical estimates from sample averages, either \textit{single path} or \textit{vine}):
\begin{align}
  % L_{\pi_{\text{old}}}{(\pi)}
  L_{\theta_{\text{old}}}{(\theta)}
=
  {\mathbb{E}}_{\bs\sim\rho^{\pi_{\theta{\text{old}}}},\ba\sim\pi_{\theta{\text{old}}}}
  \left[ \frac{\pi_{\theta}(\ba|\bs)}{\pi_{\theta{\text{old}}}(\ba|\bs)} A^{\pi_{\theta{\text{old}}}}(\bs,\ba) \right].
  \label{equ:cost-approx2}
\end{align}

% Standard policy graident methods ascend on the $\nth{1}$ order gradient, where an increase on
% $L_{\theta_{\text{old}}}{(\theta)}$
% % $L_{\pi_{\text{old}}}{(\pi)}$
% does not gaurantee an increase on
% $\eta({\pi_{\theta}})$
% % $\eta({\pi})$
% with large step sizes, due to the approximations made above.

% TRPO extends the policy improvement bound in the mixture policies setting given by \citet{kakade2002approximately} to general stochastic policies, and formulate the following optimization problem
Then it turns the soft constraint in Eq. \ref{equ:soft} into the following hard constraint problem:
\begin{align}
&\hspace{-0.1in}
\underset{\theta}{\text{minimize}}
&\hspace{-0.05in} &
{\mathbb{E}}_{\bs\sim\rho^{\pi_{\theta{\text{old}}}},\ba\sim\pi_{\theta{\text{old}}}}
\left[ \frac{\pi_{\theta}(\ba|\bs)}{\pi_{\theta{\text{old}}}(\ba|\bs)} Q^{\pi_{\theta{\text{old}}}}(\bs,\ba) \right],
\label{equ:hard1}
\\
&\hspace{-0.1in}
\text{subject to}
&\hspace{-0.05in} &
{\mathbb{E}}_{\bs\sim\rho^{\pi_{\theta{\text{old}}}}}
\left[D_{\text{KL}}(\pi_{\theta_{\text{old}}}(\cdot|\bs) || \pi_{\theta}(\cdot|\bs))\right]\leq\delta.
\label{equ:hard2}
\end{align}

% \begin{equation*}
% \begin{aligned}
% & \underset{\theta}{\text{minimize}}
% & & {\mathbb{E}}_{\bs\sim\rho^{\pi_{\theta{\text{old}}}},\ba\sim\pi_{\theta{\text{old}}}}
% \left[ \frac{\pi_{\theta}(\ba|\bs)}{\pi_{\theta{\text{old}}}(\ba|\bs)} Q^{\pi_{\theta{\text{old}}}}(\bs,\ba) \right]\\
% & \text{subject to}
% & & {\mathbb{E}}_{\bs\sim\rho^{\pi_{\theta{\text{old}}}}}
% \left[
% D_{\text{KL}}(\pi_{\theta_{\text{old}}}(\cdot|\bs) || \pi_{\theta}(\cdot|\bs))\right]
% \leq\delta.
% & &
% \end{aligned}
% \end{equation*}
where $\delta$ is a hyper parameter for the upper bound of the KL divergence between the old and the updated policy (e.g., $\delta=0.01$).
This constrained optimization problem is solved using a conjugate gradient followed by a line search;
we refer readers to \citet{schulman2015trust} for a detailed description.

\subsubsection{\textbf{PPO} \citep{schulman2017proximal}}:
Instead of reformulating a hard constraint problem as in TRPO (Eq. \ref{equ:hard1} and \ref{equ:hard2}),
PPO solves the original soft constraint optimization (Eq. \ref{equ:soft}) with \nth{1}-order SGD,
adapting $C$ according to the KL divergence.
Since it is much simpler implementation-wise compared to TRPO and gives a googd performance, PPO has become the default DRL algorithm at OpenAI.
A distributed version of PPO has also been proposed \citep{heess2017emergence}.

\subsubsection{\textbf{ACKTR} \citep{wu2017scalable}}:
The Actor Critic Kronecker-Factored Trust Region (ACKTR) is a scalable trust region natural gradient method for a actor-critic,
with the Kronecker-factored approximation to the curvature.
It is more computationally efficient than TRPO,
and is more sample efficient than those methods taking steps in the gradient direction (e.g, A2C) rather than the natural gradient direction.

\subsection{\textbf{DRL Mechanisms}}
\label{sec:drl-mechanisms}

Many useful mechanisms have also been proposed that can be added on top of the aforementioned DRL algorithms. These mechanisms generally work orthogonally with the algorithms, and some can accelerate the DRL training by a large margin. Below we list several conceptually simple yet very effective ones.

\begin{itemize}
  \item \textbf{Auxiliary Tasks} \citep{mirowski2016learning,jaderberg2016reinforcement,levine2016end,yu2018one,martin2018learning}: Uses additional supervised or unsupervised tasks (e.g., regressing depth images from color images, detecting loop closures, predicting end-effector poses) alongside the main reinforcement learning task, to compensate for the sparse supervision signals usually provided to DRL agents.
  \item \textbf{Prioritized Experience Replay} \citep{schaul2015prioritized}: Prioritizes memory replay according to \textit{td-error}; can be added to \textit{off-policy} methods.
  \item \textbf{Hindsight Experience Replay} \citep{andrychowicz2017hindsight}: Relabels the reward for collected experiences to make better use of failure trajectories, and effectively speed up \textit{off-policy} methods with \textit{binary} or \textit{sparse} reward structures.
  \item \textbf{Curriculum Learning} \citep{bengio2009curriculum,florensa2017reverse,zhang2017neural}: Presents the learning agent with progressively more complex task settings, such that it can grasp gradually more sophasticated skills.
  \item \textbf{Curiosity-driven Exploration} \citep{pathak2017curiosity}: Augments the standard external reward with internal reward measured by intrinsic motivation. % TODO: add in zheloo's ws once it's done
  \item \textbf{Asymmetric Self-replay for Exploration} \citep{sukhbaatar2017intrinsic}: Drives exploration through an automatic curricula generated via the interplay of two versions of the same agent.
  \item \textbf{Noise in Parameter Space for Exploration} \citep{fortunato2018noisy,plappert2018parameter}: Pertubates network parameters to aid exploration.
\end{itemize}

\subsection{\textbf{DRL for Navigation}}
\label{sec:drl-navigation}

Autonomous navigation is one of the essential problems and challenges in mobile robotics.
It can roughly be described as the ability of a robot to plan and follow a trajectory through the environment to reach a certain goal location without colliding with any obstacles in between.
The recent literature has seen a growing number of proposed methods that tackle the task of autonomous navigation with DRL algorithms.
Those works formulate the navigation problem as MDPs or POMDPs that first take in the sensor readings (color/depth images, laser scans, etc.) as observations and stack or augment them into states,
and then search for the optimal policy that is capable of guiding the agent to navigate to goal locations in a timely and collision-free manner.
Below we discuss several representative works in this category, that target the field of robotics.

\citet{zhu2017target} input both the first-person view and the image of the target object to the A3C model,
formulating a target-driven navigation problem based on the universal value function approximators \citep{schaul2015universal}.
The training of their model requires the features output from the pretrained \textit{ResNet}-$50$ \citep{he2016deep},
and is performed in an indoor simulator \citep{kolve2017ai2}
where each new room is regarded as a new scene for which several scene-specific layers are added as another output head of the model.
The success rate for generalizing the navigation policies to new targets one step away from the trained targets is $70\%$,
and around is $42\%$ for those that are two steps away.
For navigation tasks with optimal solutions of $17.6$ steps,
\citet{zhu2017target} achieved $210.7$ average trajectory lengths after being trained on $100$ million frames with an A3C agent.
The trained policy was able to navigate a real robot inside an office environment after being fine-tuned on images collected from the real scene.

\citet{zhang2017deep} work on a deep \textit{successor representation} formulation \citep{kulkarni2016deep,barreto2017successor} for the $Q$-value function (Eqs. \ref{equ:sr1},\ref{equ:sr2},\ref{equ:sr3}),
targeting learning representations that are transferrable between related navigation tasks.
Following the observation that most of the \textit{value-based} DRL methods, such as DQN,
usually learn a black-box function approximator for the optimal value functions,
which makes how to transfer the knowledge gained from one task to a related task unclear,
they extend on the \textit{successor feature representation} that decouples the learning of the optimal value functions into two parts,
learning task-specific reward functions, and learning task-specific features, and how those features evolve under the current task dynamics.
While this representation has been shown to work well on transferring learned policies to differently scaled reward functions and changed goals in fixed environments,
\citet{zhang2017deep} extend the formulations to cope with transferring policies to new environments.
% They deployed a network containing $3$ convolutional layers and $3$ fully connected layers and require relateively less training samples; also the training for transferring to new environments is significantly sped up).
Both experiments in a simulated 3D maze with RGB inputs and real-world robotic experiments with depth image inputs are presented.
The trained agents, either pre-trained or transferred, all achieved near-optimal performance,
validating the ability of the proposed method to transfer DRL navigation policies into new environments.

The two methods mentioned above propose to learn navigation policies without a requirement for performing localization or mapping as in the traditional planning pipelines in robotics.
They deal with navigating to different targets either by feeding the target image as input \citep{zhu2017target} or by treating it as a transfer problem \citep{zhang2017deep}.
\citet{tai2017virtual}, in contrast, propose a learning-based mapless motion planner,
under the assumption that the relative position of the target w.r.t. the robot can be obtained via cheap solutions such as \textit{wifi} or \textit{visible light} localization,
which are applicable to indoor robotic systems such as vacuum robots.
The inputs for the model is 10-dimensional laser ranges,
and the network outputs continuous steering commands after being trained via an asynchronous version of the DDPG.
Since the simulated laser ranges and the real laser readings are quite similar, the trained model is directly generalizable to indoor office environments.

\citet{mirowski2016learning} greatly improve the data efficiency and task performance of their variant of an A3C agent when learning to navigate in simulated 3D mazes, by using additional supervision signals from auxiliary tasks. In particular, the learning agent is additionally supervised by losses from depth prediction and loop closure classification. Extensive experiments are presented, validating the ability of the proposed agent to localize and to navigate between frequently changing start and goal locations.

The aforementioned methods all deal with navigation in static environments.
\citet{chen2017socially} propose a DRL based systematic solution for socially aware navigation in dynamic environments with pedestrians.
They extend a prior work \citep{chen2017decentralized}, and build a robotic system for their real-world experiment,
where a differential-drive mobile robot is mounted with a Lidar for localization and three \textit{Intel Realsense}'s for obstacle avoidance.
From the sensor readings, the speed, velocity and radius of pedestrians are estimated, from which the reward (designed based on social norms) is calculated.
Read robotic experiments show that the proposed method is capable of navigating agents at human walking speed in a dynamic environment with many pedestrians.

\citet{long2017towards} deal with decentralized multi-agent collision avoidance with PPO. They supervised the agents with a well-shaped reward function, and test the algorithm under extensive simulated scenarios.

There is also a growing trend in the recent literature to incorporate traditional Simultaneous Localization and Mapping (SLAM) \citep{thrun2005probabilistic} procedures,
either partially or fully and embedded internally or externally, into DRL network architectures,
with the intention to cultivate more sophisticated navigation capabilities in DRL agents \citep{stachenfeld2017hippocampus,kanitscheider2017training}. Below we review the most representative robotics works in this promising direction.

\citet{gupta2017cognitive} train a Cognitive Mapping and Planning (CMP) model with \textit{DAGGer},
which is an \textit{imitation learning} algorithm that we will talk about in Sec. \ref{sec:imi-bc}.
Although it dose not use DRL for training the navigation policies, we feel it fits best into this part of the discussion.
CMP takes in the first-person view RGB images and applies egomotion to an internal mapper module,
to encourage an egocentric multi-scale map representation to emerge out of the training process.
Planning is done on this egocentric map utilizing Value Iteration Networks (VIN) \citep{tamar2016value}.
Also trained with \textit{DAGGer}, \citet{gupta2017unifying} unify map-based spatial reasoning and path planning.
Given the images and poses of several reference points and the starting point, as well as the pose for the goal,
their proposed method is able to navigate toward the agent the desired goal location.

\citet{zhang2017neural} proposed \textit{Neural SLAM},
based on the \textit{Neural Map} proposed by \citet{parisotto2017neural},
where the \textit{Neural Turing Machine} (NTM) is deployed for the DRL agent to interact with.
More specifically, \textit{Neural SLAM} embeds the motion prediction step and the measurement update step of traditional SLAM into the network architecture,
biasing the write/read operations in the NTM towards SLAM operations,
and treats the external memory as an internal representation of the environment for the learning agent.
The whole architecture is then trained via A3C,
and experiments show that the embedded structures are able to encourage the evolution of cognitive mapping capabilities of the agent,
during the course of its exploration through the environment.

\citet{khan2017memory} design a network architecture that contains three components: a convolution network for feature extraction, a planner module to pre-plan in the environment, and a controller module that learns to selectively store past information that could be useful for planning.

\citet{bruce2017one} propose a method that enables zero-shot transfer for a mobile robot to learn to navigate to a fixed goal in an environment with variations unseen during the training.
They introduce \textit{interactive replay}, in which a rough world model is built from a single traversal of the environment.
The agent is then able to interact with this world model to generate a large number of diverse trajectories,
which can substantially reduce the number of real experiences needed.

\citet{chaplot2018active} introduce a network structure mimicking the Bayesian filtering process with a specially designed perception model.
It takes as input the agent's observation, and outputs a likelihood map, inside which the belief is propagated through time following the classic filtering process used for localization in robotics \citep{thrun2006probabilistic}.

Inspired by graph-based SLAM algorithms \citep{thrun2005probabilistic,kummerle2011g},
\citet{parisotto2018global} embed the global pose optimization into the network architecture design for their \textit{Neural Graph Optimizer},
which is composed of a local pose estimation model, a pose selection module and a graph optimization process.
\citet{savinov2018semiparametric} introduce a memory architecture, \textit{Semi-parametric Topological Memory}, for navigation in unseen environments.
It contains a non-parametric graph with nodes representing locations in the environment, and a parametric deep network retrieving nodes from the graph based on observations.

\subsection{\textbf{DRL for Manipulation}}
\label{sec:drl-manipulation}

In terms of manipulation, the tasks being considered for evaluating DRL algorithms are more standardized in the recent literature \citep{lillicrap2015continuous,schulman2015trust,mnih2016asynchronous,heess2017emergence,wu2017scalable}.
Most of such works benchmark the proposed algorithms on standard tasks,
including \textit{reaching}, \textit{pushing}, \textit{pick-and-place}, etc.,
using the \textit{MuJoCo} simulator \citep{todorov2012mujoco}.
Below we focus on the works that are presented with real robotic experiments.

%\hl{@lei check again all the drl papers that have real robot manipulation experiments}
\citet{gu2017deep} propose an asynchronous version of NAF.
Taking in the low dimensional states as inputs (joint angles, end effector poses, as well as their time derivatives, and the pose of the target),
in addition to well-shaped reward signals, it allows the robot to learn a real-world door opening task in about $2.5$ hours in a completely end-to-end manner,
achieving a $100\%$ success rate.

% Considering the highly model-based character of manipulation task, it is natural to estimate the dynamic model in related tasks through model-based reinforcement learning.
% It can significantly improve the sampling efficiency.
\citet{levine2016end} successfully train deep visuomotor policies with GPS, a model-based approach.
Their proposed visuomotor policy network takes as input monocular RGB images and passes them through several convolutional layers and a spatial soft argmax layer,
which are then concatenated with the robot configurations (joint angles, end effector poses).
These representations are then passed through several fully connected layers and used to predict the corresponding motor torques.
Various experiments on a \textit{PR2} robot (with a $7$-DOF arm)
such as hanging a coat hanger on a clothes rack, inserting a block into a shape sorting cube, or screwing on a bottle cap
have demonstrated to validate the effectiveness of the approach.
This method, however, requires a known and fully observed state space, which could limit its potential use cases.

% Followed the model-based reinforcement learning method, like GPS mentioned above,
Model-based DRL methods are also utilized by
\citet{finn2016deep} and \citet{tzeng2015towards}, learning useful state representations for generating successful control policies. % deployed the trained reinforcement learning policies in real world directly through visual state learning and domain constriains.
\citet{fu2016one} proposed one-shot learning of manipulation skills through model-based reinforcement learning by leveraging the neural network priors as a dynamic model.
Learning dexterous manipulation skills with multi-fingered hands,
for which model-based \citep{kumar2016optimal, gupta2016learning} and model-free \citep{popov2017data} DRL algorithms have been proposed and demonstrated in real robotic experiments,
is quite challenging.

%\citep{sermanet2017time} \hl{using LQR, should we talk about this?}

While many works have carefully designed their reward structure to guide reinforcement learning,
\citet{martin2018learning} propose a method to speed up learning from only binary or sparse rewards,
under the observation that well-shaped rewards can often bias the learned control policy into potentially suboptimal directions.
In contrast when only sparse reward signals are provided to the agent, the learner can discover novel and potentially preferable solutions.
To achieve this, alongside the policy learning for the main task, \citet{martin2018learning} learn policies (which they refer to as intentions) for a set of semantically grounded auxiliary tasks,
whose supervision signals can be easily obtained by the activation of certain sensors.
Then a scheduling policy is learned to sequence the intention-policies.
Their proposed algorithm is able to learn to solve challenging manipulation tasks from scratch,
such as stacking two blocks into a tower or cleaning up a desk by putting objects desk into a box with a lid that can be opened,
with a $9$-DOF robot arm. Moreover, in their real-world experiments, a single robot arm learns a lifting task in about $10$ hours.

\subsection{\textbf{The Reality Gap: From Simulation to the Real World}} \label{sec:reality-gap}

Although DRL offers a general framework for agents to learn high-dimensional control policies,
it typically requires several millions of training samples.
This makes it infeasible to train DRL agents directly in real-world scenarios,
since real robotic control experiences are relatively expensive to obtain.
As a consequence, DRL algorithms are generally trained in simulated environments,
then transferred to the real world and deployed onto real robotic systems.
This brings about the problem of the \textit{reality gap}, which refers to the discrepancies in lighting conditions,
noise patterns, textures, etc., between synthetic renderings and real-world sensory readings. The \textit{reality gap} imposes major challenges for generalizing the DRL policies trained in simulation to real scenarios.

Since the problem of the \textit{reality gap} is most severe in the visual domain,
in that the aforementioned discrepancies are most significant between rendered color images and real color camera readings,
some robotics works have proposed to circumvent this problem by using other input modalities whose domain variations are less distinct,
such as depth images \citep{zhang2017deep} or laser ranges \citep{tai2017virtual}.
However,
%since color camera is a relatively cheap and \hl{useful @lei think of another word} solution for robotics perception,
bridging the \textit{reality gap} in the visual domain is of great importance and remains one of the focuses of recent works.
Below, we review methods that deal with the \textit{reality gap} for visual control.

\subsubsection{\textbf{Domain Adaptation}}
In the visual domain, \textit{domain adaptation} can also be referred to as \textit{image-to-image translation},
which focuses on translating images from a source domain to a target domains,
It can be considered as the method of choice in the recent literature to tackle the \textit{reality gap} for visual control.
The \textit{domain confusion loss}, as proposed by \citet{tzeng2014deep}, is another solution that learns a semantically meaningful and domain invariant representation.
However, minimizing the \textit{domain confusion loss} requires that the data from both the source and the target domain are available from the beginning of the whole learning pipeline,
which might not be as flexible in the robotics context.

In the following, we first formalize the \textit{domain adaptation} problem,
then continue to introduce several of the most general methods that require the least human intervention and are most directly applicable to robotics control tasks.

Consider visual data sources from two domains: $\bX$ (e.g., synthetic images rendered by a simulator; $\bx\sim p_{\text{sim}}$, where $p_{\text{sim}}$ represents the simulated data distribution) and $\bY$ (e.g., real sensory readings from the onboard color camera of a mobile robot; $\by\sim p_{\text{real}}$, where $p_{\text{real}}$ represents the distribution of the real color image readings). As we have just discussed, DRL agents are typically trained in the synthetic domain $\bX$, then deployed onto real robotic platforms to perform control tasks in the real-world domain $\bY$. \textit{Domain adaptation} methods aim to learn a mapping between these two domains.

% \begin{itemize}
\textbf{GANs}: Most of the \textit{domain adaptation} works are based on Generative Adversarial Networks (GANs) \citep{goodfellow2014generative,radford2015unsupervised,arjovsky2017wasserstein}.
When learning a GAN model, a generator $G$ and a discriminator $D$ are trained in an adversarial manner.
In the context of \textit{domain adaptation} for visual inputs,
the generator $G$ takes images from the source domain,
and tries to generate output images matching those from the target domain,
while the discriminator $D$ learns to tell the generated target images and the real target images apart.

\textbf{\textit{CycleGAN}} \citep{zhu2017unpaired}: Zhu et al. propose one of the most popular unsupervised \textit{domain adaptation} methods in the recent literature,
\citet{zhu2017unpaired} proposed a simple yet very effective formulation that does not require paired data from the two domains of interest.
Observing that the mapping from the source domain to the target domain, $G_{\bY}: \bX\rightarrow\bY$, is highly under-constrained,
\textit{CycleGAN} proposes to add a cycle-consistent loss to enforce that a reverse mapping from the target domain back to the source domain exists: $G_{\bX}: \bY\rightarrow\bX$.
  More formally, \textit{CycleGAN} learns two generative models to map between domains $\bX$ and $\bY$: $G_{\bY}$ with its discriminator $D_{\bY}$, and $G_{\bX}$ with its discriminator $D_{\bX}$, by training two GANs simultaneously:
\begin{align}
    &\mathcal{L}_{\text{GAN}_{\bY}}(G_{\bY},D_{\bY};\bX,\bY)=
\\&\hspace{0.3in}
    \mathbb{E}_{\by} \left[ \log D_{\bY}(\by) \right]
+
    \mathbb{E}_{\bx} \left[ \log (1-D_{\bY}(G_{\bY}(\bx))) \right],
\label{equ:gan-y-loss}
\nonumber\\
    &\mathcal{L}_{\text{GAN}_{\bX}}(G_{\bX},D_{\bX};\bY,\bX)=
\\&\hspace{0.3in}
    \mathbb{E}_{\bx} \left[ \log D_{\bX}(\bx) \right]
+
    \mathbb{E}_{\by} \left[ \log (1-D_{\bX}(G_{\bX}(\by))) \right],
\nonumber
\label{equ:gan-x-loss}
\end{align}
on top of which two sets of \textit{cycle consistency loss} are added to constrain the two mappings:
\begin{align}
    \mathcal{L}_{\text{cyc}_{\bY}}(G_{\bX},G_{\bY};\bY)
&=
    \mathbb{E}_{\by}\left[ \left|\left| G_{\bY}(G_{\bX}(\by))-\by \right|\right|_{1} \right],
\\
    \mathcal{L}_{\text{cyc}_{\bX}}(G_{\bY},G_{\bX};\bX)
&=
    \mathbb{E}_{\bx}\left[ \left|\left| G_{\bX}(G_{\bY}(\bx))-\bx \right|\right|_{1} \right].
    \label{equ:cycle-loss}
\end{align}

The full objective of \textit{CycleGAN} then adds up to ($\lambda$ denotes the weighting of the \textit{cycle consistency loss})
\begin{align}
    \mathcal{L}(G_{\bY} & , G_{\bX},D_{\bY},D_{\bX};\bX,\bY)
=\nonumber
\\\nonumber&\hspace{0.18in}
    \mathcal{L}_{\text{GAN}_{\bY}}(G_{\bY},D_{\bY};\bX,\bY)
\\\nonumber&+
    \mathcal{L}_{\text{GAN}_{\bX}}(G_{\bX},D_{\bX};\bY,\bX)
\\\nonumber&+
    \lambda_{\text{cyc}}
    \mathcal{L}_{\text{cyc}_{\bY}}(G_{\bX},G_{\bY};\bY)
\\&+
    \lambda_{\text{cyc}}
    \mathcal{L}_{\text{cyc}_{\bX}}(G_{\bY},G_{\bX};\bX),
\end{align}
which corresponds to the following optimization problem:
\begin{align}
    G_{\bY}^*, G_{\bX}^*
&=
    \arg\min_{G_{\bY},G_{\bX}}\max_{D_{\bY},D_{\bX}}\mathcal{L}(G_{\bY},G_{\bX},D_{\bY},D_{\bX}).
\label{equ:full-obj}
\end{align}

This conceptually simple method works surprisingly well in practice,
especially in domains with relatively few semantic types (e.g., when the source domain images contain only horses and background, and the target domain images contain only zebras and background),
where it is less challenging for the algorithm to find the matching semantics between the two domains (e.g., horse $\leftrightarrow$ zebra).
However, the results of \textit{CycleGAN} on translating between more complex data distributions containing many more semantic types,
such as between urban street scenario images and their corresponding semantic labels, are not as satisfactory,
in that the generators often permute the labels for some semantics.

\textbf{\textit{CyCADA}} \citep{hoffman2017cycada}: The semantic consistency loss proposed in \textit{CyCADA} offers a good solution to learning the mapping between more complex data distributions with relatively more semantic types.
To be more specific, in \textit{CyCADA}, a semantic segmentation network $f$ is first trained in the domain where semantic labels are available (e.g., $f_{\bX}$ for the synthetic domain $\bX$).
(This is applicable for the \textit{domain adaptation} between the simulated domain $\bX$ and the real-world domain $\bY$ in the context of robotics,
since many recent robotics simulators provide the ground truth semantic maps of the rendered images, while the labels for the real images are expensive to obtain.)
Then this semantic segmentation network is used to constrain the semantic consistency between the input and the translated output images of the generators:

\begin{align}
    % \mathcal{L}_{\text{sem}_{X}}(G_Y, X;f_X)&
    \mathcal{L}_{\text{sem}_{\bY}}(G_{\bY};\bX,f_{\bX})&
=
    % \text{CrossEnt}(f_{X}(\bx),f_{X}(G_{Y}(\bx)))
    \text{CrossEnt}(f_{\bX}(\bX),f_{\bX}(G_{\bY}(\bX))),
\\
    % \mathcal{L}_{\text{sem}_{X}}(G_X, Y;f_X)&
    \mathcal{L}_{\text{sem}_{\bX}}(G_{\bX};\bY,f_{\bX})&
=
    % \text{CrossEnt}(f_{X}(\by),f_{X}(G_{X}(\by)))
    \text{CrossEnt}(f_{\bX}(\bY),f_{\bX}(G_{\bX}(\bY))),
\end{align}
where $\text{CrossEnt}(S_{\bX},f_{\bX}(\bX))$ represents the cross-entropy loss between the semantics of $X$ predicted by the pretrained semantic segmentation network $f_{\bX}$ and the ground truth label $S_{\bX}$. The semantic consistency losses are then added to the \textit{CycleGAN} objective (Eq. \ref{equ:full-obj}).
% \end{itemize}

\subsubsection{\textbf{Domain Adaptation for Visual DRL Policies}}
  % \item \textbf{Domain Adaptation for Visual DRL Policies}
% for those reality-gap paragrphas
While many extensions and variants have been proposed for \textit{image-to-image translation} in the computer vision literature,
here we focus on those \textit{domain adaptation} methods that specifically itarget transferring DRL control policies from simulation to real scenarios.

% manipulation
For manipulation tasks, \citet{bousmalis2017using} deal with the \textit{reality gap} by adapting synthetic images to the realistic domain before feeding them into the DRL policy network during the training phase.
However, the additional adaptation step required for every training iteration could significantly slow down the whole learning pipeline.
\citet{tobin2017domain} proposed to randomise the lighting conditions, viewing angles and textures of objects during the training phase of the DRL policies in simulation, in the hope that
after being exposed to enough variations,
the learned model can naturally generalize to real-world scenarios.
However,
this method can only be applied to simulators where such randomization can be easily achieved at a low cost,
which is not the case for most of the popular robotic simulators.
Moreover, there is no guarantee that for a random real-world scenario, its visual modality can be covered by the randomized simulations.
Similarly, randomizing the dynamics of the simulator during training has also been proposed \citep{peng2017sim} to bridge the \textit{reality gap}.
\cite{rusu2016sim} propose to progressively adapt the learned deep features and representations from the synthetic domain to the real-world domain.
However, this method still requires going through an expensive DRL control policy training phase (although this procedure can be greatly sped up by initial training in the simulator) for each new visually different real-world scene.

% navigation
The aforementioned methods realize \textit{domain adaptation} via the \textit{sim-to-real} direction,
meaning that they either translate the synthetic images to the real-world domain during the training of DRL policies,
or adapt the deep features of the simulated domain to those of the realistic domain. However, the DRL policy learning and the adaptation of the policies are entangled in this line of methods.

The recently proposed model of \textit{VR Goggles} \citep{zhang2018vr} decouples the two components of policy learning and policy adaptation,
by tackling the \textit{reality gap} from the \textit{real-to-sim} direction,
which requires no extra transfer steps during the expensive training of DRL policies.
Specifically, the \textit{VR Goggles} deal with the \textit{reality gap} only during the actual deployment phase,
by translating real-world sensor reading streams back to the simulated domain,
so as to adapt the unseen or unfamiliar characteristics of the real scenes to the synthetic features,
which the agent has already learned well how to deal with, to \textit{make the robot feel at home}.
To constrain the consistency between the generated subsequent frames,
a \textit{shift loss} is added to the optimization objective,
which is inspired by the \textit{artistic style transfer for videos} literature \citep{ruder2017artistic}.
This method is validated in transferring DRL navigation policies, which could be considered more challenging than manipulation tasks,
since the environments the navigation agents operate in are typically at much larger scales than the confined workspace of manipulators.

Both results of outdoor and indoor scene adaptation have been presented.
For the outdoor experiment, the synthetic data is collected from the \textit{CARLA} simulator \citep{dosovitskiy2017carla}，
which provides the ground truth semantic labels, and the real world data is gathered from the \textit{RobotCar} dataset \citep{RobotCarDatasetIJRR}.
The semantic consistency loss is added for the outdoor scenario, with a semantic segmentation network trained using the \textit{DeepLab} model \citep{chen2016deeplab}.
The semantic consistency is critical for outdoor scenes containing various semantic types,
without such a constraint, permutation of semantics occurs.
It is also critical for situations where the model fails to generate a virtual car at the position at which there is a real car in the real image
(This kind of performance is as reported by \citet{yang2018unsupervised} whithout constraining the semantic consistency),
which could potentially lead to accidents in self-driving scenarios.

For indoor scenes, the semantic loss is not added,
as the simulated domain \textit{Gazebo} \citep{koenig2004design} does not provide ground truth labels,
and also the real scene, which is a real office environment,
contains relatively fewer semantic types.
A real robot (\textit{Turtlebot3 Waffle}) is deployed in the office environment and feed its sensor readings (captured by a \textit{RealSense R200 camera}) to the \textit{VR Goggles} model. The translated \textit{Gazebo} images are then fed to the DRL policy network to give control commands.
The \textit{VR Goggles} offer a lightweight and flexible solution for transferring DRL visual control policies from simulation to the real world,
and should also be applicable to manipulation tasks.

% The aforementioned \textit{domain adaptation} methods all target at manipulation tasks. In terms of dealing with the \textit{reality gap} for navigation policies, however, there exist far less literature. This is partially due to the fact that, which imposes more challenges. Still, as mentioned before, several works circumvent this problem by using modalities that are more consistent between the synthetic and realistic domains \citep{zhang2017deep,tai2017virtual}.

% However, the semantic consistency before and after the image translation should be critical for the method, since failing to generate a virtual car where there is a real car in the real image could lead to vital accidents in self-driving (these kind of performance is reported in the method proposed by \citep{yang2018unsupervised}, which does not utilize the semantic consistency loss), for which the method semantic consistency could be of great importance. The \textit{VR Goggles} as proposed in \citep{zhang2018vr}, incorporated the semantic loss, as well as a \textit{shift loss} .

% \end{itemize}

\subsection{\textbf{Simulation Platforms}} \label{sec:sim}

% \multirow{2}{*}{Approach} & \multirow{2}{*}{Sensor} & Deep & Dataset&Real-time & \multirow{2}{*}{Effect} & \multirow{2}{*}{Application} &Training&\multirow{2}{*}{Precision}\\
% \begin{table}[h!]
% \small
% \centering
% \caption{Categorization of simulated environments for robots.\label{tab:T1}}
% \resizebox{\columnwidth}{!}
% {\begin{tabular}{ccccc}
% \toprule
% Simulator                         & Sensor Observation  & Headless mode & Task  \\
% \midrule
% Gazebo \citep{koenig2004design}    &  Sensor Plugins        &  Yes  & General Robot Tasks  \\
% Vrep \citep{rohmer2013v}           &  Sensor Plugins        &  Yes  & General Robot Tasks \\
% Airsim \citep{airsim2017fsr}       &  Depth/Color/Semantic Labels      &  No   & Outdoor Navigation \\
% Carla \citep{dosovitskiy2017carla} &  Depth/Color/Semantic Labels      &  No   & Autonomous Driving  \\
% Torcs                              &  Color                 &  Yes  & Autonomous Driving  \\
% GTA5                               &  Color                 &  No   & Autonomous Driving  \\
% Minos \citep{savva2017minos}       &  Reconfigurable        &  No   & Indoor Navigation    \\
% House3D \citep{wu2018building}     &  Depth/Color/Semantic Labels      &  No   & Indoor Navigation  \\
% %Deep \cite{tassa2018deepmind} &                      &
% \bottomrule
% \end{tabular}}
% \end{table}
As mentioned before, DRL algorithms, at their current state,
are in general not sample efficient enough to be directly trained on real robotic platforms.
Thus robotics simulators are utilized for the initial training of DRL policies.
Here we review several of the most widely used simulation platforms that are suitable for DRL training.

% In the following we first summerize the common pitfalls, \hl{todo}
% then continue to review several most widely-used simulation platforms that are suitable for DRL training.

\begin{table*}[h!]
% \big
\centering
\caption{Robotic Simulators.\label{tab:T1}
}
\resizebox{\textwidth}{!}
{\begin{tabular}{cccc}
\toprule
% Simulator                          & Modalities     & Headless Mode Enabled & Framerate & Target Use Case  \\
% \midrule
% Gazebo \citep{koenig2004design}    &  Sensor Plugins        &  Yes  & 10s+FPS & General Purposes  \\
% Vrep \citep{rohmer2013v}           &  Sensor Plugins        &  Yes  & 10s+FPS & General Purposes \\
% Airsim \citep{airsim2017fsr}       &  Depth/Color/Semantics &  No   & 20s+FPS & Autonomous Driving \\
% Carla \citep{dosovitskiy2017carla} &  Depth/Color/Semantics &  Yes   & 30s+FPS & Autonomous Driving  \\
% Torcs \citep{you2017virtual}       &  Color/Semantics       &  Yes  & 100s+FPS & Autonomous Driving  \\
% AI2-Thor \citep{kolve2017ai2}      &  Color                 &  -    & 100s+FPS    & Indoor Navigation   \\
% Minos \citep{savva2017minos}       &  Depth/Color/Semantics &  No   & 100s+FPS & Indoor Navigation    \\
% House3D \citep{wu2018building}     &  Depth/Color/Semantics &  No   & 100s+FPS & Indoor Navigation  \\
Simulator                          & Modalities     & Framerate & Target Use Case  \\
\midrule
Gazebo \citep{koenig2004design}    &  Sensor Plugins          & 10s+FPS & General Purposes  \\
Vrep \citep{rohmer2013v}           &  Sensor Plugins          & 10s+FPS & General Purposes \\
Airsim \citep{airsim2017fsr}       &  Depth/Color/Semantics   & 20s+FPS & Autonomous Driving \\
Carla \citep{dosovitskiy2017carla} &  Depth/Color/Semantics   & 30s+FPS & Autonomous Driving  \\
Torcs \citep{you2017virtual}       &  Color/Semantics         & 100s+FPS & Autonomous Driving  \\
AI2-Thor \citep{kolve2017ai2}      &  Color                     & 100s+FPS    & Indoor Navigation   \\
Minos \citep{savva2017minos}       &  Depth/Color/Semantics    & 100s+FPS & Indoor Navigation    \\
House3D \citep{wu2018building}     &  Depth/Color/Semantics   & 600s+FPS & Indoor Navigation  \\
\bottomrule
\end{tabular}}
\end{table*}

%\hl{@lei, rewrite this part and add points that I didn't think of}
We summarize the most commonly used simulators in Table \ref{tab:T1},
listing their available sensor observation types and their target use cases.

\section{\textbf{Imitation Learning}}
\label{sec:imitaaa}
%\hl{@lei TODO should also have a big picture here}

DRL offers a formulation for control skills acquisition.
However, relying on learning from trial and error, DRL methods typically require a significant amount of system interaction time.
Also, a carefully designed well-shaped reward structure is necessary to guide the search of optimal policies,
which can often be non-trivial in complex scenarios.

Imitation learning, as an alternative to learning control policies,
guides the policy search, not by hand-designed reward signals,
but by providing the learning agent with experts' demonstrations \citep{bagnell2015invitation}.
It offers a paradigm for agents to learn successful policies in fields where people can easily demonstrate the desired behavior but find it difficult to hand program or hardcode the correct cost or reward function.
This is especially useful for humanoid robots or manipulators with high degrees of freedom.

Perhaps the most simple method of \textit{imitation learning} is addressing it as a standard \textit{supervised learning} problem.
But as we have discussed,
%in the beginning of Sec. \ref{sec:control},
as a method for learning policies to make sequential control decisions,
\textit{imitation learning} cannot be conducted effectively by directly applying the classical \textit{supervised learning} approaches.
We emphasize the most critical distinctions below:

% \begin{itemize}
\textbf{Independent VS.\ Compounding Errors}: Standard \textit{supervised learning} assumes that the predictions made by the learning agents do not affect the environment in which they operate;
hence the data distribution they are to encounter is assumed to be the same as what they have experienced.
However, although the learning errors are independent for each sample in \textit{supervised learning},
they are \textit{compounded} in \textit{imitation learning}.
This is due to the fact that the standard \textit{supervised learning} algorithms are only expected to do well over samples that are drawn from the same distribution as they have been trained on.
This i.i.d. assumption, however,is badly violated in \textit{imitaiton learning}, in which an early error could potentially \textit{cascade} to a sequence of mistakes, carried out by the control decisions that are made sequentially by the learning agent.

\textbf{Single-Timestep VS.\ Multi-Timestep Decisions}: \textit{Supervised learning} agents are only capable of learning \textit{reactive policies},
since they completely ignore the temporal dependence between subsequent decisions,
which leads to \textit{myopic} strategies.
In contrast, for making informative decisions, classical planning approaches in robotics reason far into the future (but often require sophisticatedly designed cost functions).
Also, a naive imitation of the experts' demonstrations often misses the true learning objective:
instead of copying the demonstrated behaviors given by the experts,
the actual goal of \textit{imitation learning} is in some cases quite different and inexplicitly optimized in the demonstrations,
such as to increase the success rate of accomplishing a specific task, to minimize the probability of colliding with obstacles,
or to minimize the total travel cost.
% \end{itemize}

In the following, we proceed by going through the three most common approaches of \textit{imitation learning},
which address the above issues from different perspectives,
and introduce representative works in robotics for each.

\subsection{\textbf{Behavior Cloning}}
\label{sec:imi-bc}

Behavior cloning tackles the problem of \textit{imitation learning} in a supervised manner,
by directly learning the mapping between the input observations and their corresponding actions, which are given by the expert policy.
This simple formulation can give a satisfactory performance when there is enough training data,
but will lead to \textit{compounding errors}, as we have just discussed.
One of the most well-known algorithms to compensate for this is \textit{DAGGer} (in which \textit{DAGG} stands for \textit{Data AGGregation}) \citep{ross2011reduction},
which interleaves execution and learning.
To be more specific, in the $i\text{th}$ iteration of \textit{DAGGer},
the current learned policy $\pi_{i-1}$ will be executed to collect experiences.
Then those newly recorded observations will be relabeled by the expert policy $\pi_{\text{E}}$.
These corrected new experiences $D_{i}$ will be added to the existing dataset $D$,
on which a new policy $\pi_{i}$ is trained.
This interaction between execution and learning halts the error compounding and bounds the expected error to that in the standard \textit{supervised learning} setting.

% Although \textit{DAGGer} is a relatively practical solution, the interaction learning procedure still is not always possible to be carried out in the context of robotics. so there are ... direct supervised learing approach
% circumvent this problem by learning robust embeddings

Due to its simple formulation, behavior cloning has been widely studied and applied in robotics control problems.

We start with the literature in the field of navigation and self-driving imitation.
\citet{bojarski2016end} learn a direct mapping from raw first-person view color images to steering commands,
on a training dataset collected by driving on a wide variety of roads and in diverse weather and lighting conditions,
which in total adds up to $72$ hours of driving data.
\citet{tai2016deep} drive an indoor mobile robot autonomously through a dataset based on joystick commands from human demonstrator.
A depth visual image is taken as the only input in their implementation.
\citet{giusti2016machine} train a deep network to determine actions that can keep a quadrotor on a trail,
by learning on single monocular images collected from the robot's perspective.
Eight hours of video is captured using three GoPros mounted on the head of a hiker,
with one pointing to the left, one to the right, and one straight ahead.
The optimal actions for the collected images can then be easily labeled;
 e.g., the quadrotor should turn right when facing an image collected from the left-facing camera.
\citet{codevilla2017end} observes that the pure \textit{behavior cloning} assumption could break under certain situations,
such as when a driver approaches an intersection.
The driver's subsequent actions cannot be fully explained by the observations,
since they are additionally affected by the driver's internal throughouts, such as the intended destination.
To address this, a conditional \textit{imitation learning} method is proposed to additionally constrain the \textit{imitation learning} additionally on a representation of the expert's intention,
so as to resolve the ambiguity in the perceptuomotor mapping.
Both simulated and real-world experiments are conducted,
in which the synthetic dataset is collected in the simulated self-driving environment \textit{Carla} \citep{dosovitskiy2017carla} and a real-world dataset from remote controlling a robotic truck in a residential area,
each of which contains two hours of driving time.

In terms of \textit{imitation learning} for manipulation, a recently proposed line three works presents and improves on
\textit{one-shot imitation learning}: from taking low dimensional states and expert action pairs as demonstrations \citep{duan2017one},
to learning from demonstrations of raw visual images paired with actions \citep{finn2017one},
and finally arriving at the current state of learning from human demonstration videos without labeled actions \citep{yu2018one}.
Below we discuss these methods in more detail.

\citet{duan2017one} present the imitation agent with pairs of demonstrations for each iteration during training, in which the network takes as input the firsth demonstration and a state sampled from the second demonstration.
The network is then trained using behavior cloning losses to predict the corresponding action of that sampled state.
The concrete example used in their problem setting is a distribution of block stacking tasks,
in which the goal is to control a robot arm to stack various numbers of cubic blocks into configurations specified by the user.
Each observation is a list of the relative positions of the blocks w.r.t. the gripper, and information indicating whether the gripper is open or closed.
Several architectural designs, such as temporal dropout and convolutions, neighborhood attention,
are incorporated into their training pipeline to cope with variable-dimensional and potentially long sequences of inputs.
In their experiments, the performance of pure \textit{behavior cloning} achieves the same level of performance as training with \textit{DAGGer},
suggesting that at least for this specific block-stacking task, the interactive supervision in \textit{DAGGer} might not necessarily lead to a performance gain.

\citet{finn2017one} and \citet{yu2018one} both extend the \textit{Model-Agnostic Meta-Learning} (MAML) method \citep{finn2017model},
which we will briefly review here before proceeding.
The objective of MAML is to learn a model, such that,
after being trained on a variety of learning tasks,
it is able to learn to solve new tasks with only a small number of training samples. Formally, this model of interest is denoted as $f_{\theta}$ with weights $\theta$, and the meta-learning is considered over a distribution of tasks $p(\mathcal{T})$. The model parameters will be updated from $\theta$ to $\theta'_i$, when adapting to a new task $\mathcal{T}_i$. This update is performed using gradient descent on task $\mathcal{T}_i$:
\begin{align}
    \theta'_{i}
&=
    \theta - \alpha\nabla_{\theta}\mathcal{L}_{\mathcal{T}_i}(f_{\theta}),
\end{align}
where $\alpha$ denotes a step size, and $\mathcal{L}$ represents a behavior cloning loss function (e.g., mean squared error for continuous actions, cross-entropy loss for discrete actions).
After the updated $\theta'_{i}$ is obtained, its performance is optimized w.r.t. $\theta$ across tasks sampled from $p(\mathcal{T})$, leading to the following meta-learning objective:
\begin{align}
    \min\sum_{\mathcal{T}_i\sim p(\mathcal{T})} \mathcal{L}_{\mathcal{T}_i}(f_{\theta'_{i}})
&=
    \sum_{\mathcal{T}_i\sim p(\mathcal{T})} \mathcal{L}_{\mathcal{T}_i}(f_{\theta-\alpha\nabla_{\theta}\mathcal{L}_{\mathcal{T}_i}(f_{\theta})}),
\end{align}
which is performed via SGD such that $\theta$ is updated as follows:
\begin{align}
    \theta
&\leftarrow
    \theta - \beta\nabla_{\theta}\sum_{\mathcal{T}_i\sim p(\mathcal{T})} \mathcal{L}_{\mathcal{T}_i}(f_{\theta'_{i}}),
\end{align}
where $\beta$ is the meta step size.

Here, the meta-optimization is performed over $\theta$,
while the loss is computed using the updated parameters $\theta'$.
This objective will help to find model parameters that are sensitive to changes in the task,
such that small changes in the parameters could lead to large improvements in the performance on any task sampled from $p(\mathcal{T})$.

Based on the formulation of MAML, \citet{finn2017one} learn policies that can be quickly adapted to new tasks using a single demonstration.
Here each observation input into the model contains a color image from the robot's perspective,
and the robot configurations (joint angles, end-effector poses).
While both \citet{duan2017one} and \citet{finn2017one} use only robot demonstrations throughout training and testing,
\citet{yu2018one} is able to cope with domain shift by learning from both robot and human demonstrations,
in which the human demonstrations are not labeled with expert actions.
After meta-learning,
the proposed method is capable of learning from human videos. To cope with the unlabelled human demonstrations,
an adaptation loss function $\mathcal{L}_{\psi}$ is learned alongside the meta-learning objective.
During training, human demonstrations are used to compute the updated policy parameters $\theta'_{i}$ with the gradients calculated using $\mathcal{L}_{\psi}$.
Then the performance of $\theta'_{i}$ is evaluated using the behavior cloning loss to update both $\theta$ and $\psi$.
Note that all the robot demonstrations are collected via teleoperation \citep{zhang2017deepimi}.

A recent work from \citet{eitel2017learning} introduces a model that is able to propose push actions based on over-segmented RGB-D images,
in order to separate unknown objects in cluttered environments.

\subsection{\textbf{Inverse Reinforcement Learning}}
\label{sec:imi-irl}

Inverse reinforcement learning (IRL) frames \textit{imitation learning} as solutions to MDPs,
thus reducing the problem of learning to the problem of recovering a utility function that makes the demonstrated behavior (near-)optimal.
After this utility function is obtained, reinforcement learning procedures can be performed on top to search for optimal policies.
A representative IRL method, the Maximum Entropy IRL \citep{ziebart2008maximum}, fits a cost function from a family of functions $\mathcal{C}$ to optimize the following objective:

\begin{align}
    \argmax_{c\in\mathcal{C}} \left( \min_{\pi\in\prod} -H(\pi)+\mathbf{E}_{\pi} \left[ c(\bs,\ba) \right] \right) - \mathbf{E}_{\pi^{\text{E}}} \left[ c(\bs,\ba) \right],
\end{align}
where $\prod$ denotes the family of policies.

In robotics, the formulation of IRL offers an efficient solution for learning policies for socially compliant navigation \citep{okal2016learning, pfeiffer2016predicting, kretzschmar2016socially},
where the agent needs to not only avoid collisions with static obstacles but also to behave in a socially compliant manner.
Thus, the underlying cost function is non-trivial to hand-design, but the hebaviors are easy to demonstrate.

\citet{wulfmeier2015maximum} extends Maximum Entropy IRL under the deep learning context,
utilizing fully convolutional neural networks as the approximator for learning the reward function.
The proposed algorithm is successfully deployed for learning the cost map in urban environments,
from a dataset of driving behaviors demonstrated by human experts.

\subsection{\textbf{Generative Adversarial Imitation Learning}}
\label{sec:imi-gail}

The learning process of IRL can be indirect and slow.
Inspired by Generative Adversarial Networks (GANs) \citep{goodfellow2014generative},
\citet{ho2016generative} propose Generative Adversarial Imitation Learning (GAIL),
which surpasses the intermediate step of learning a reward function
and is capable of directly learning a policy from expert demonstrations.
To be more specific, in the GAIL model,
a generator $\pi_{\theta}$ with parameters $\theta$ is trained to generate state-action ($\mathcal{S}\times\mathcal{A}$)
pairs matching those of the expert demonstrations, while the discriminator $D_{\omega}$ learns to tell apart the generated policy $\pi_{\theta}$ from the expert (demonstrated) policy $\pi^{\text{E}}$. The GAIL optimization objective is defined as follows:
\begin{align}
    \mathbf{E}_{\pi_{\theta}}\left[ \log(D(\bs,\ba)) \right]
+
    \mathbf{E}_{\pi^{\text{E}}}\left[ \log(1-D(\bs,\ba)) \right]
-
    \lambda H(\pi_{\theta}),
    \label{equ:gail}
\end{align}
where $H(\pi_{\theta})$ denotes the causal entropy. The training of GAIL interleaves between updating parameters $\omega$ of the discriminator $D_{\omega}$ to maximize Eq. \ref{equ:gail},
and utilizing DRL techniques such as TRPO to minimize Eq. \ref{equ:gail} w.r.t. the parameters $\theta$ of the generator $\pi_{\theta}$.
The scores given out by the discriminator for the generated experiences are regarded as costs for those state-action pairs for TRPO.
Several extensions of GAIL have also been proposed \citep{baram2016model,wang2017robust}.

In the field of navigation, \citet{li2017inferring} successfully apply GAIL in simulated autonomous vehicle navigation scenarios with raw visual input.
\citet{tai2018social} learn a socially compliant navigation policy through GAIL, based on raw depth input,
and demonstrate the learned behaviors in real robotics experiments.

For manipulation, \citet{stadie2017third} extend the GAIL formulation with ideas from domain confusion loss \citep{tzeng2014deep}, and successfully utilize it to train agents to imitate third-person demonstrations, by learning a domain-agnostic representation of the agent's observations.

\section{Challenges and Open Research Questions}

Utilizing deep learning techniques for learning control for robotics tasks has shown great potential.
Yet, there still remain many challenges for scaling up and stabilizing the aforementioned algorithms to meet the requirements of operating robotics systems in real-world applications. We list the critical challenges and the corresponding future research directions.

\begin{itemize}
  \item \textbf{Sample Efficiency}:
  Gathering experiences by interacting with the environment for \textit{deep reinforcement learning},
  or collecting expert demonstrations for \textit{imitation learning},
  are both expensive procedures,
  in terms of executing control commands on real robotics systems.
  Thus, designing sample efficient algorithms is of critical importance.
  \item \textbf{Strong Real-time Requirements}:
  A single forward pass of very deep networks with millions of parameters could be relatively slow
  if not equipped with special computation hardware and might not meet the real-time requirement for controlling real robotics systems.
  Learning compact representations for dexterous policies is preferable.
  \item \textbf{Safety Concerns}:
  Real robotics systems, such as mobile robots, quadrotors or self-driving cars,
  are expected to operate in environments that could be highly dynamic and potentially dangerous.
  Also, unlike a wrong prediction from a perception model,
  which would not cascade or affect the physical robotic systems or the environment, a single false output might lead to a serious accident.
  Thus, attention should be paid to include practical considerations to bound the uncertainty of the possible outcomes when deploying control policies on real autonomous systems.
  \item \textbf{Stability, Robustness and Interpretability}:
  DRL algorithms could be relatively unstable,
  and their performance might deviate a lot between configurations that only differ slightly from each other \citep{henderson2017deep}.
  To overcome this problem, gaining more insight into the learned representations and the policies,
  could be beneficial for detecting adversarial scenarios to prevent robotic systems from safety threats.
  \item \textbf{Lifelong Learning}:
  The visual appearance of the environments that autonomous agents operate in cab vary dramatically during different seasons of the year,
  or even different times of day,
  which could hinder the performance of the learned control policies.
  Hence the ability of continuing to learn to adapt to environmental changes as well as preserving the solutions to the already experienced scenarios could be of critical value.
  \item \textbf{Generalization Between Tasks}: Most of the aforementioned algorithms are designed to excel in a particular task,
  which is not ideal, as intelligent robotic systems are expected to be capable of carrying out a set of tasks, with a minimal total training time for all considered tasks.
\end{itemize}

In contrast, with the rapid development of deep learning, several research directions are gaining much attention for robotics.

\begin{itemize}
  \item \textbf{Unifying Reinforcement Learning and Imitation Learning}:
  Several recent works \citep{vevcerik2017leveraging,nair2017overcoming,gao2018reinforcement,zhu2018reinforcement} have introduced algorithms that unify reinforcement learning and imitation learning
  such that the learning agent can benefit from both expert demonstrations and interactions with the environment.
  This setup can be beneficial for learning control,
  as pure DRL algorithms are, in general, relatively expensive to train,
  while learning purely by imitating demonstrated behaviors can restrict or bias the control policy in potentially suboptimal directions.
  Thus, the method of using demonstrations to kick-start the policy learning,
  then applying DRL methods to adjust the learned policy, can potentially lead to advanced performance.
  \item \textbf{Meta-learning}: \citet{finn2017model} and \citet{nichol2018reptile} propose methods that can effectively lead the policy search to find parameters that can be adapted to give good performance on a new task with only a small number of training examples of the new task.
  Such formulations could be very beneficial, and have the potential to learn universal and robust policies.
\end{itemize}

\section{Conclusion}  \label{sec:conclu}

In this paper, we give a brief overview of deep learning solutions for robotics control tasks,
focusing mainly on \textit{deep reinforcement learning} and \textit{imitation learning} algorithms.
We mainly introduce the formulations for each learning paradigm and the corresponding representative works in robotics.
Finally, We discuss the challenges and potential future research directions.

\ifCLASSOPTIONcaptionsoff
\newpage
\fi



%\bibliographystyle{IEEEtran}
\bibliographystyle{SageH}
\bibliography{tai2018survey}

\begin{thebibliography}{127}
\providecommand{\natexlab}[1]{#1}
\providecommand{\url}[1]{\texttt{#1}}
\providecommand{\urlprefix}{URL }
\expandafter\ifx\csname urlstyle\endcsname\relax
  \providecommand{\doi}[1]{DOI:\discretionary{}{}{}#1}\else
  \providecommand{\doi}{DOI:\discretionary{}{}{}\begingroup
  \urlstyle{rm}\Url}\fi

\bibitem[{Andrychowicz et~al.(2017)Andrychowicz, Crow, Ray, Schneider, Fong,
  Welinder, McGrew, Tobin, Abbeel and Zaremba}]{andrychowicz2017hindsight}
Andrychowicz M, Crow D, Ray A, Schneider J, Fong R, Welinder P, McGrew B, Tobin
  J, Abbeel OP and Zaremba W (2017) Hindsight experience replay.
\newblock In: \emph{Advances in Neural Information Processing Systems}. pp.
  5055--5065.

\bibitem[{Arjovsky et~al.(2017)Arjovsky, Chintala and
  Bottou}]{arjovsky2017wasserstein}
Arjovsky M, Chintala S and Bottou L (2017) Wasserstein gan.
\newblock \emph{arXiv preprint arXiv:1701.07875} .

\bibitem[{Bagnell(2015)}]{bagnell2015invitation}
Bagnell JA (2015) An invitation to imitation.
\newblock Technical report, CARNEGIE-MELLON UNIV PITTSBURGH PA ROBOTICS INST.

\bibitem[{Baram et~al.(2016)Baram, Anschel and Mannor}]{baram2016model}
Baram N, Anschel O and Mannor S (2016) Model-based adversarial imitation
  learning.
\newblock \emph{arXiv preprint arXiv:1612.02179} .

\bibitem[{Barreto et~al.(2017)Barreto, Dabney, Munos, Hunt, Schaul, Silver and
  van Hasselt}]{barreto2017successor}
Barreto A, Dabney W, Munos R, Hunt JJ, Schaul T, Silver D and van Hasselt HP
  (2017) Successor features for transfer in reinforcement learning.
\newblock In: \emph{Advances in Neural Information Processing Systems}. pp.
  4058--4068.

\bibitem[{Bengio et~al.(2009)Bengio, Louradour, Collobert and
  Weston}]{bengio2009curriculum}
Bengio Y, Louradour J, Collobert R and Weston J (2009) Curriculum learning.
\newblock In: \emph{Proceedings of the 26th annual international conference on
  machine learning}. ACM, pp. 41--48.

\bibitem[{Bojarski et~al.(2016)Bojarski, Del~Testa, Dworakowski, Firner, Flepp,
  Goyal, Jackel, Monfort, Muller, Zhang et~al.}]{bojarski2016end}
Bojarski M, Del~Testa D, Dworakowski D, Firner B, Flepp B, Goyal P, Jackel LD,
  Monfort M, Muller U, Zhang J et~al. (2016) End to end learning for
  self-driving cars.
\newblock \emph{arXiv preprint arXiv:1604.07316} .

\bibitem[{Bousmalis et~al.(2017)Bousmalis, Irpan, Wohlhart, Bai, Kelcey,
  Kalakrishnan, Downs, Ibarz, Pastor, Konolige et~al.}]{bousmalis2017using}
Bousmalis K, Irpan A, Wohlhart P, Bai Y, Kelcey M, Kalakrishnan M, Downs L,
  Ibarz J, Pastor P, Konolige K et~al. (2017) Using simulation and domain
  adaptation to improve efficiency of deep robotic grasping.
\newblock \emph{arXiv preprint arXiv:1709.07857} .

\bibitem[{Bruce et~al.(2017)Bruce, S{\"u}nderhauf, Mirowski, Hadsell and
  Milford}]{bruce2017one}
Bruce J, S{\"u}nderhauf N, Mirowski P, Hadsell R and Milford M (2017) One-shot
  reinforcement learning for robot navigation with interactive replay.
\newblock \emph{arXiv preprint arXiv:1711.10137} .

\bibitem[{Chaplot et~al.(2018)Chaplot, Parisotto and
  Salakhutdinov}]{chaplot2018active}
Chaplot DS, Parisotto E and Salakhutdinov R (2018) Active neural localization.
\newblock \emph{arXiv preprint arXiv:1801.08214} .

\bibitem[{Chen et~al.(2016)Chen, Papandreou, Kokkinos, Murphy and
  Yuille}]{chen2016deeplab}
Chen LC, Papandreou G, Kokkinos I, Murphy K and Yuille AL (2016) Deeplab:
  Semantic image segmentation with deep convolutional nets, atrous convolution,
  and fully connected crfs.
\newblock \emph{arXiv preprint arXiv:1606.00915} .

\bibitem[{Chen et~al.(2017{\natexlab{a}})Chen, Everett, Liu and
  How}]{chen2017socially}
Chen YF, Everett M, Liu M and How JP (2017{\natexlab{a}}) Socially aware motion
  planning with deep reinforcement learning.
\newblock \emph{arXiv preprint arXiv:1703.08862} .

\bibitem[{Chen et~al.(2017{\natexlab{b}})Chen, Liu, Everett and
  How}]{chen2017decentralized}
Chen YF, Liu M, Everett M and How JP (2017{\natexlab{b}}) Decentralized
  non-communicating multiagent collision avoidance with deep reinforcement
  learning.
\newblock In: \emph{Robotics and Automation (ICRA), 2017 IEEE International
  Conference on}. IEEE, pp. 285--292.

\bibitem[{Codevilla et~al.(2017)Codevilla, M{\"u}ller, Dosovitskiy, L{\'o}pez
  and Koltun}]{codevilla2017end}
Codevilla F, M{\"u}ller M, Dosovitskiy A, L{\'o}pez A and Koltun V (2017)
  End-to-end driving via conditional imitation learning.
\newblock \emph{arXiv preprint arXiv:1710.02410} .

\bibitem[{Dayan(1993)}]{dayan1993improving}
Dayan P (1993) Improving generalization for temporal difference learning: The
  successor representation.
\newblock \emph{Neural Computation} 5(4): 613--624.

\bibitem[{Deisenroth et~al.(2013)Deisenroth, Neumann, Peters
  et~al.}]{deisenroth2013survey}
Deisenroth MP, Neumann G, Peters J et~al. (2013) A survey on policy search for
  robotics.
\newblock \emph{Foundations and Trends{\textregistered} in Robotics} 2(1--2):
  1--142.

\bibitem[{Deng(2014)}]{deng2014tutorial}
Deng L (2014) A tutorial survey of architectures, algorithms, and applications
  for deep learning.
\newblock \emph{APSIPA Transactions on Signal and Information Processing} 3.

\bibitem[{Dosovitskiy et~al.(2017)Dosovitskiy, Ros, Codevilla, L{\'o}pez and
  Koltun}]{dosovitskiy2017carla}
Dosovitskiy A, Ros G, Codevilla F, L{\'o}pez A and Koltun V (2017) Carla: An
  open urban driving simulator.
\newblock \emph{arXiv preprint arXiv:1711.03938} .

\bibitem[{Duan et~al.(2017)Duan, Andrychowicz, Stadie, Ho, Schneider,
  Sutskever, Abbeel and Zaremba}]{duan2017one}
Duan Y, Andrychowicz M, Stadie B, Ho OJ, Schneider J, Sutskever I, Abbeel P and
  Zaremba W (2017) One-shot imitation learning.
\newblock In: \emph{Advances in neural information processing systems}. pp.
  1087--1098.

\bibitem[{Eitel et~al.(2017)Eitel, Hauff and Burgard}]{eitel2017learning}
Eitel A, Hauff N and Burgard W (2017) Learning to singulate objects using a
  push proposal network.
\newblock In: \emph{Proc.~of the International Symposium on Robotics Research
  (ISRR)}. Puerto Varas, Chile.

\bibitem[{Finn et~al.(2017{\natexlab{a}})Finn, Abbeel and
  Levine}]{finn2017model}
Finn C, Abbeel P and Levine S (2017{\natexlab{a}}) Model-agnostic meta-learning
  for fast adaptation of deep networks.
\newblock \emph{arXiv preprint arXiv:1703.03400} .

\bibitem[{Finn et~al.(2016)Finn, Tan, Duan, Darrell, Levine and
  Abbeel}]{finn2016deep}
Finn C, Tan XY, Duan Y, Darrell T, Levine S and Abbeel P (2016) Deep spatial
  autoencoders for visuomotor learning.
\newblock In: \emph{Robotics and Automation (ICRA), 2016 IEEE International
  Conference on}. IEEE, pp. 512--519.

\bibitem[{Finn et~al.(2017{\natexlab{b}})Finn, Yu, Zhang, Abbeel and
  Levine}]{finn2017one}
Finn C, Yu T, Zhang T, Abbeel P and Levine S (2017{\natexlab{b}}) One-shot
  visual imitation learning via meta-learning.
\newblock \emph{arXiv preprint arXiv:1709.04905} .

\bibitem[{Florensa et~al.(2017)Florensa, Held, Wulfmeier and
  Abbeel}]{florensa2017reverse}
Florensa C, Held D, Wulfmeier M and Abbeel P (2017) Reverse curriculum
  generation for reinforcement learning.
\newblock \emph{arXiv preprint arXiv:1707.05300} .

\bibitem[{Fortunato et~al.(2018)Fortunato, Azar, Piot, Menick, Osband, Graves,
  Mnih, Munos, Hassabis, Pietquin et~al.}]{fortunato2018noisy}
Fortunato M, Azar MG, Piot B, Menick J, Osband I, Graves A, Mnih V, Munos R,
  Hassabis D, Pietquin O et~al. (2018) Noisy networks for exploration.
\newblock In: \emph{International Conference on Learning Representations}.
\newblock \urlprefix\url{https://openreview.net/forum?id=rywHCPkAW}.

\bibitem[{Fu et~al.(2016)Fu, Levine and Abbeel}]{fu2016one}
Fu J, Levine S and Abbeel P (2016) One-shot learning of manipulation skills
  with online dynamics adaptation and neural network priors.
\newblock In: \emph{Intelligent Robots and Systems (IROS), 2016 IEEE/RSJ
  International Conference on}. IEEE, pp. 4019--4026.

\bibitem[{Fu et~al.(2005)Fu, Glover and April}]{fu2005simulation}
Fu MC, Glover FW and April J (2005) Simulation optimization: a review, new
  developments, and applications.
\newblock In: \emph{Proceedings of the 37th conference on Winter simulation}.
  Winter Simulation Conference, pp. 83--95.

\bibitem[{Gao et~al.(2018)Gao, Xu, Lin, Yu, Levine and
  Darrell}]{gao2018reinforcement}
Gao Y, Xu HH, Lin J, Yu F, Levine S and Darrell T (2018) Reinforcement learning
  from imperfect demonstrations .

\bibitem[{Giusti et~al.(2016)Giusti, Guzzi, Cire{\c{s}}an, He, Rodr{\'\i}guez,
  Fontana, Faessler, Forster, Schmidhuber, Di~Caro et~al.}]{giusti2016machine}
Giusti A, Guzzi J, Cire{\c{s}}an DC, He FL, Rodr{\'\i}guez JP, Fontana F,
  Faessler M, Forster C, Schmidhuber J, Di~Caro G et~al. (2016) A machine
  learning approach to visual perception of forest trails for mobile robots.
\newblock \emph{IEEE Robotics and Automation Letters} 1(2): 661--667.

\bibitem[{Goodfellow et~al.(2016)Goodfellow, Bengio and
  Courville}]{Goodfellow-et-al-2016}
Goodfellow I, Bengio Y and Courville A (2016) \emph{Deep Learning}.
\newblock MIT Press.
\newblock \url{http://www.deeplearningbook.org}.

\bibitem[{Goodfellow et~al.(2014)Goodfellow, Pouget-Abadie, Mirza, Xu,
  Warde-Farley, Ozair, Courville and Bengio}]{goodfellow2014generative}
Goodfellow I, Pouget-Abadie J, Mirza M, Xu B, Warde-Farley D, Ozair S,
  Courville A and Bengio Y (2014) Generative adversarial nets.
\newblock In: \emph{Advances in neural information processing systems}. pp.
  2672--2680.

\bibitem[{Gu et~al.(2017)Gu, Holly, Lillicrap and Levine}]{gu2017deep}
Gu S, Holly E, Lillicrap T and Levine S (2017) Deep reinforcement learning for
  robotic manipulation with asynchronous off-policy updates.
\newblock In: \emph{Robotics and Automation (ICRA), 2017 IEEE International
  Conference on}. IEEE, pp. 3389--3396.

\bibitem[{Gu et~al.(2016)Gu, Lillicrap, Sutskever and
  Levine}]{gu2016continuous}
Gu S, Lillicrap T, Sutskever I and Levine S (2016) Continuous deep q-learning
  with model-based acceleration.
\newblock In: \emph{International Conference on Machine Learning (ICML-16)}.

\bibitem[{Guo et~al.(2016)Guo, Liu, Oerlemans, Lao, Wu and Lew}]{guo2016deep}
Guo Y, Liu Y, Oerlemans A, Lao S, Wu S and Lew MS (2016) Deep learning for
  visual understanding: A review.
\newblock \emph{Neurocomputing} 187: 27--48.

\bibitem[{Gupta et~al.(2016)Gupta, Eppner, Levine and
  Abbeel}]{gupta2016learning}
Gupta A, Eppner C, Levine S and Abbeel P (2016) Learning dexterous manipulation
  for a soft robotic hand from human demonstrations.
\newblock In: \emph{Intelligent Robots and Systems (IROS), 2016 IEEE/RSJ
  International Conference on}. IEEE, pp. 3786--3793.

\bibitem[{Gupta et~al.(2017{\natexlab{a}})Gupta, Davidson, Levine, Sukthankar
  and Malik}]{gupta2017cognitive}
Gupta S, Davidson J, Levine S, Sukthankar R and Malik J (2017{\natexlab{a}})
  Cognitive mapping and planning for visual navigation.
\newblock \emph{arXiv preprint arXiv:1702.03920} .

\bibitem[{Gupta et~al.(2017{\natexlab{b}})Gupta, Fouhey, Levine and
  Malik}]{gupta2017unifying}
Gupta S, Fouhey D, Levine S and Malik J (2017{\natexlab{b}}) Unifying map and
  landmark based representations for visual navigation.
\newblock \emph{arXiv preprint arXiv:1712.08125} .

\bibitem[{He et~al.(2016)He, Zhang, Ren and Sun}]{he2016deep}
He K, Zhang X, Ren S and Sun J (2016) Deep residual learning for image
  recognition.
\newblock In: \emph{Proceedings of the IEEE conference on computer vision and
  pattern recognition}. pp. 770--778.

\bibitem[{Heess et~al.(2017)Heess, Sriram, Lemmon, Merel, Wayne, Tassa, Erez,
  Wang, Eslami, Riedmiller et~al.}]{heess2017emergence}
Heess N, Sriram S, Lemmon J, Merel J, Wayne G, Tassa Y, Erez T, Wang Z, Eslami
  A, Riedmiller M et~al. (2017) Emergence of locomotion behaviours in rich
  environments.
\newblock \emph{arXiv preprint arXiv:1707.02286} .

\bibitem[{Henderson et~al.(2017)Henderson, Islam, Bachman, Pineau, Precup and
  Meger}]{henderson2017deep}
Henderson P, Islam R, Bachman P, Pineau J, Precup D and Meger D (2017) Deep
  reinforcement learning that matters.
\newblock \emph{arXiv preprint arXiv:1709.06560} .

\bibitem[{Ho and Ermon(2016)}]{ho2016generative}
Ho J and Ermon S (2016) Generative adversarial imitation learning.
\newblock In: \emph{Advances in Neural Information Processing Systems}. pp.
  4565--4573.

\bibitem[{Hoffman et~al.(2017)Hoffman, Tzeng, Park, Zhu, Isola, Saenko, Efros
  and Darrell}]{hoffman2017cycada}
Hoffman J, Tzeng E, Park T, Zhu JY, Isola P, Saenko K, Efros AA and Darrell T
  (2017) Cycada: Cycle-consistent adversarial domain adaptation.
\newblock \emph{arXiv preprint arXiv:1711.03213} .

\bibitem[{Huang et~al.(2017)Huang, Liu, Weinberger and van~der
  Maaten}]{huang2017densely}
Huang G, Liu Z, Weinberger KQ and van~der Maaten L (2017) Densely connected
  convolutional networks.
\newblock In: \emph{Proceedings of the IEEE conference on computer vision and
  pattern recognition}, volume~1. p.~3.

\bibitem[{Jaderberg et~al.(2016)Jaderberg, Mnih, Czarnecki, Schaul, Leibo,
  Silver and Kavukcuoglu}]{jaderberg2016reinforcement}
Jaderberg M, Mnih V, Czarnecki WM, Schaul T, Leibo JZ, Silver D and Kavukcuoglu
  K (2016) Reinforcement learning with unsupervised auxiliary tasks.
\newblock \emph{arXiv preprint arXiv:1611.05397} .

\bibitem[{Kakade and Langford(2002)}]{kakade2002approximately}
Kakade S and Langford J (2002) Approximately optimal approximate reinforcement
  learning.
\newblock In: \emph{ICML}, volume~2. pp. 267--274.

\bibitem[{Kalweit and Boedecker(2017)}]{pmlr-v78-kalweit17a}
Kalweit G and Boedecker J (2017) Uncertainty-driven imagination for continuous
  deep reinforcement learning.
\newblock In: Levine S, Vanhoucke V and Goldberg K (eds.) \emph{Proceedings of
  the 1st Annual Conference on Robot Learning}, \emph{Proceedings of Machine
  Learning Research}, volume~78. PMLR, pp. 195--206.
\newblock \urlprefix\url{http://proceedings.mlr.press/v78/kalweit17a.html}.

\bibitem[{Kanitscheider and Fiete(2017)}]{kanitscheider2017training}
Kanitscheider I and Fiete I (2017) Training recurrent networks to generate
  hypotheses about how the brain solves hard navigation problems.
\newblock In: \emph{Advances in Neural Information Processing Systems}. pp.
  4532--4541.

\bibitem[{Khan et~al.(2017)Khan, Zhang, Atanasov, Karydis, Kumar and
  Lee}]{khan2017memory}
Khan A, Zhang C, Atanasov N, Karydis K, Kumar V and Lee DD (2017) Memory
  augmented control networks.
\newblock \emph{arXiv preprint arXiv:1709.05706} .

\bibitem[{Koenig et~al.(2004)Koenig, A and Howard}]{koenig2004design}
Koenig N, A B and Howard A (2004) Design and use paradigms for gazebo, an
  open-source multi-robot simulator.
\newblock In: \emph{Intelligent Robots and Systems, 2004.(IROS 2004).
  Proceedings. 2004 IEEE/RSJ International Conference on}, volume~3. IEEE, pp.
  2149--2154.

\bibitem[{Kolve et~al.(2017)Kolve, Mottaghi, Gordon, Zhu, Gupta and
  Farhadi}]{kolve2017ai2}
Kolve E, Mottaghi R, Gordon D, Zhu Y, Gupta A and Farhadi A (2017) Ai2-thor: An
  interactive 3d environment for visual ai.
\newblock \emph{arXiv preprint arXiv:1712.05474} .

\bibitem[{Kretzschmar et~al.(2016)Kretzschmar, Spies, Sprunk and
  Burgard}]{kretzschmar2016socially}
Kretzschmar H, Spies M, Sprunk C and Burgard W (2016) Socially compliant mobile
  robot navigation via inverse reinforcement learning.
\newblock \emph{The International Journal of Robotics Research} 35(11):
  1289--1307.

\bibitem[{Krizhevsky et~al.(2012)Krizhevsky, Sutskever and
  Hinton}]{krizhevsky2012imagenet}
Krizhevsky A, Sutskever I and Hinton GE (2012) Imagenet classification with
  deep convolutional neural networks.
\newblock In: \emph{Advances in neural information processing systems}. pp.
  1097--1105.

\bibitem[{Kulkarni et~al.(2016)Kulkarni, Saeedi, Gautam and
  Gershman}]{kulkarni2016deep}
Kulkarni TD, Saeedi A, Gautam S and Gershman SJ (2016) Deep successor
  reinforcement learning.
\newblock \emph{arXiv preprint arXiv:1606.02396} .

\bibitem[{Kumar et~al.(2016)Kumar, Todorov and Levine}]{kumar2016optimal}
Kumar V, Todorov E and Levine S (2016) Optimal control with learned local
  models: Application to dexterous manipulation.
\newblock In: \emph{Robotics and Automation (ICRA), 2016 IEEE International
  Conference on}. IEEE, pp. 378--383.

\bibitem[{K{\"u}mmerle et~al.(2011)K{\"u}mmerle, Grisetti, Strasdat, Konolige
  and Burgard}]{kummerle2011g}
K{\"u}mmerle R, Grisetti G, Strasdat H, Konolige K and Burgard W (2011) g 2 o:
  A general framework for graph optimization.
\newblock In: \emph{Robotics and Automation (ICRA), 2011 IEEE International
  Conference on}. IEEE, pp. 3607--3613.

\bibitem[{Levine et~al.(2016)Levine, Finn, Darrell and Abbeel}]{levine2016end}
Levine S, Finn C, Darrell T and Abbeel P (2016) End-to-end training of deep
  visuomotor policies.
\newblock \emph{The Journal of Machine Learning Research} 17(1): 1334--1373.

\bibitem[{Levine and Koltun(2013)}]{levine2013guided}
Levine S and Koltun V (2013) Guided policy search.
\newblock In: \emph{ICML (3)}. pp. 1--9.

\bibitem[{Li et~al.(2017)Li, Song and Ermon}]{li2017inferring}
Li Y, Song J and Ermon S (2017) Inferring the latent structure of human
  decision-making from raw visual inputs.
\newblock \emph{arXiv preprint arXiv:1703.08840} .

\bibitem[{Lillicrap et~al.(2015)Lillicrap, Hunt, Pritzel, Heess, Erez, Tassa,
  Silver and Wierstra}]{lillicrap2015continuous}
Lillicrap TP, Hunt JJ, Pritzel A, Heess N, Erez T, Tassa Y, Silver D and
  Wierstra D (2015) Continuous control with deep reinforcement learning.
\newblock In: \emph{Proceedings of the International Conference on Learning
  Representations (ICLR)}.

\bibitem[{Long et~al.(2015)Long, Shelhamer and Darrell}]{long2015fully}
Long J, Shelhamer E and Darrell T (2015) Fully convolutional networks for
  semantic segmentation.
\newblock In: \emph{Proceedings of the IEEE Conference on Computer Vision and
  Pattern Recognition}. pp. 3431--3440.

\bibitem[{Long et~al.(2017)Long, Fan, Liao, Liu, Zhang and
  Pan}]{long2017towards}
Long P, Fan T, Liao X, Liu W, Zhang H and Pan J (2017) Towards optimally
  decentralized multi-robot collision avoidance via deep reinforcement
  learning.
\newblock \emph{arXiv preprint arXiv:1709.10082} .

\bibitem[{Maddern et~al.(2017)Maddern, Pascoe, Linegar and
  Newman}]{RobotCarDatasetIJRR}
Maddern W, Pascoe G, Linegar C and Newman P (2017) {1 Year, 1000km: The Oxford
  RobotCar Dataset}.
\newblock \emph{The International Journal of Robotics Research (IJRR)} 36(1):
  3--15.
\newblock \doi{10.1177/0278364916679498}.
\newblock \urlprefix\url{http://dx.doi.org/10.1177/0278364916679498}.

\bibitem[{Mirowski et~al.(2016)Mirowski, Pascanu, Viola, Soyer, Ballard,
  Banino, Denil, Goroshin, Sifre, Kavukcuoglu et~al.}]{mirowski2016learning}
Mirowski P, Pascanu R, Viola F, Soyer H, Ballard A, Banino A, Denil M, Goroshin
  R, Sifre L, Kavukcuoglu K et~al. (2016) Learning to navigate in complex
  environments.
\newblock \emph{arXiv preprint arXiv:1611.03673} .

\bibitem[{Mnih et~al.(2016)Mnih, Badia, Mirza, Graves, Lillicrap, Harley,
  Silver and Kavukcuoglu}]{mnih2016asynchronous}
Mnih V, Badia AP, Mirza M, Graves A, Lillicrap TP, Harley T, Silver D and
  Kavukcuoglu K (2016) Asynchronous methods for deep reinforcement learning.
\newblock \emph{arXiv preprint arXiv:1602.01783} .

\bibitem[{Mnih et~al.(2015)Mnih, Kavukcuoglu, Silver, Rusu, Veness, Bellemare,
  Graves, Riedmiller, Fidjeland, Ostrovski et~al.}]{mnih2015human}
Mnih V, Kavukcuoglu K, Silver D, Rusu AA, Veness J, Bellemare MG, Graves A,
  Riedmiller M, Fidjeland AK, Ostrovski G et~al. (2015) Human-level control
  through deep reinforcement learning.
\newblock \emph{Nature} 518(7540): 529--533.

\bibitem[{Nair et~al.(2017)Nair, McGrew, Andrychowicz, Zaremba and
  Abbeel}]{nair2017overcoming}
Nair A, McGrew B, Andrychowicz M, Zaremba W and Abbeel P (2017) Overcoming
  exploration in reinforcement learning with demonstrations.
\newblock \emph{arXiv preprint arXiv:1709.10089} .

\bibitem[{Nichol and Schulman(2018)}]{nichol2018reptile}
Nichol A and Schulman J (2018) Reptile: a scalable metalearning algorithm.
\newblock \emph{arXiv preprint arXiv:1803.02999} .

\bibitem[{Okal and Arras(2016)}]{okal2016learning}
Okal B and Arras KO (2016) Learning socially normative robot navigation
  behaviors with bayesian inverse reinforcement learning.
\newblock In: \emph{ICRA}. pp. 2889--2895.

\bibitem[{Parisotto et~al.(2018)Parisotto, Chaplot, Zhang and
  Salakhutdinov}]{parisotto2018global}
Parisotto E, Chaplot DS, Zhang J and Salakhutdinov R (2018) Global pose
  estimation with an attention-based recurrent network.
\newblock \emph{arXiv preprint arXiv:1802.06857} .

\bibitem[{Parisotto and Salakhutdinov(2017)}]{parisotto2017neural}
Parisotto E and Salakhutdinov R (2017) Neural map: Structured memory for deep
  reinforcement learning.
\newblock \emph{arXiv preprint arXiv:1702.08360} .

\bibitem[{Pathak et~al.(2017)Pathak, Agrawal, Efros and
  Darrell}]{pathak2017curiosity}
Pathak D, Agrawal P, Efros AA and Darrell T (2017) Curiosity-driven exploration
  by self-supervised prediction.
\newblock In: \emph{International Conference on Machine Learning (ICML)},
  volume 2017.

\bibitem[{Peng et~al.(2017)Peng, Andrychowicz, Zaremba and
  Abbeel}]{peng2017sim}
Peng XB, Andrychowicz M, Zaremba W and Abbeel P (2017) Sim-to-real transfer of
  robotic control with dynamics randomization.
\newblock \emph{arXiv preprint arXiv:1710.06537} .

\bibitem[{Pfeiffer et~al.(2016)Pfeiffer, Schwesinger, Sommer, Galceran and
  Siegwart}]{pfeiffer2016predicting}
Pfeiffer M, Schwesinger U, Sommer H, Galceran E and Siegwart R (2016)
  Predicting actions to act predictably: Cooperative partial motion planning
  with maximum entropy models.
\newblock In: \emph{2016 IEEE/RSJ International Conference on Intelligent
  Robots and Systems (IROS)}. pp. 2096--2101.

\bibitem[{Plappert et~al.(2018)Plappert, Houthooft, Dhariwal, Sidor, Chen,
  Chen, Asfour, Abbeel and Andrychowicz}]{plappert2018parameter}
Plappert M, Houthooft R, Dhariwal P, Sidor S, Chen RY, Chen X, Asfour T, Abbeel
  P and Andrychowicz M (2018) Parameter space noise for exploration.
\newblock In: \emph{International Conference on Learning Representations}.
\newblock \urlprefix\url{https://openreview.net/forum?id=ByBAl2eAZ}.

\bibitem[{Popov et~al.(2017)Popov, Heess, Lillicrap, Hafner, Barth-Maron,
  Vecerik, Lampe, Tassa, Erez and Riedmiller}]{popov2017data}
Popov I, Heess N, Lillicrap T, Hafner R, Barth-Maron G, Vecerik M, Lampe T,
  Tassa Y, Erez T and Riedmiller M (2017) Data-efficient deep reinforcement
  learning for dexterous manipulation.
\newblock \emph{arXiv preprint arXiv:1704.03073} .

\bibitem[{Radford et~al.(2015)Radford, Metz and
  Chintala}]{radford2015unsupervised}
Radford A, Metz L and Chintala S (2015) Unsupervised representation learning
  with deep convolutional generative adversarial networks.
\newblock \emph{arXiv preprint arXiv:1511.06434} .

\bibitem[{Riedmiller et~al.(2018)Riedmiller, Hafner, Lampe, Neunert, Degrave,
  Van~de Wiele, Heess and Springenberg}]{martin2018learning}
Riedmiller M, Hafner R, Lampe T, Neunert M, Degrave J, Van~de Wiele V Tom
  adn~Mnih, Heess N and Springenberg JT (2018) Learning by playing - solving
  sparse reward tasks from scratch.
\newblock \emph{arXiv preprint arXiv:1802.10567} .

\bibitem[{Rohmer et~al.(2013)Rohmer, Singh and Freese}]{rohmer2013v}
Rohmer E, Singh SP and Freese M (2013) V-rep: A versatile and scalable robot
  simulation framework.
\newblock In: \emph{Intelligent Robots and Systems (IROS), 2013 IEEE/RSJ
  International Conference on}. IEEE, pp. 1321--1326.

\bibitem[{Ross et~al.(2011)Ross, Gordon and Bagnell}]{ross2011reduction}
Ross S, Gordon G and Bagnell D (2011) A reduction of imitation learning and
  structured prediction to no-regret online learning.
\newblock In: \emph{Proceedings of the fourteenth international conference on
  artificial intelligence and statistics}. pp. 627--635.

\bibitem[{Ruder et~al.(2017)Ruder, Dosovitskiy and Brox}]{ruder2017artistic}
Ruder M, Dosovitskiy A and Brox T (2017) Artistic style transfer for videos and
  spherical images.
\newblock \emph{arXiv preprint arXiv:1708.04538} .

\bibitem[{Rusu et~al.(2017)Rusu, Ve{\v{c}}er{\'\i}k, Roth{\"o}rl, Heess,
  Pascanu and Hadsell}]{rusu2016sim}
Rusu AA, Ve{\v{c}}er{\'\i}k M, Roth{\"o}rl T, Heess N, Pascanu R and Hadsell R
  (2017) Sim-to-real robot learning from pixels with progressive nets.
\newblock In: \emph{Conference on Robot Learning}. pp. 262--270.

\bibitem[{Savinov et~al.(2018)Savinov, Dosovitskiy and
  Koltun}]{savinov2018semiparametric}
Savinov N, Dosovitskiy A and Koltun V (2018) Semi-parametric topological memory
  for navigation.
\newblock In: \emph{International Conference on Learning Representations}.
\newblock \urlprefix\url{https://openreview.net/forum?id=SygwwGbRW}.

\bibitem[{Savva et~al.(2017)Savva, Chang, Dosovitskiy, Funkhouser and
  Koltun}]{savva2017minos}
Savva M, Chang AX, Dosovitskiy A, Funkhouser T and Koltun V (2017) Minos:
  Multimodal indoor simulator for navigation in complex environments.
\newblock \emph{arXiv preprint arXiv:1712.03931} .

\bibitem[{Schaul et~al.(2015{\natexlab{a}})Schaul, Horgan, Gregor and
  Silver}]{schaul2015universal}
Schaul T, Horgan D, Gregor K and Silver D (2015{\natexlab{a}}) Universal value
  function approximators.
\newblock In: \emph{International Conference on Machine Learning}. pp.
  1312--1320.

\bibitem[{Schaul et~al.(2015{\natexlab{b}})Schaul, Quan, Antonoglou and
  Silver}]{schaul2015prioritized}
Schaul T, Quan J, Antonoglou I and Silver D (2015{\natexlab{b}}) Prioritized
  experience replay.
\newblock \emph{arXiv preprint arXiv:1511.05952} .

\bibitem[{Schmidhuber(2015)}]{schmidhuber2015deep}
Schmidhuber J (2015) Deep learning in neural networks: An overview.
\newblock \emph{Neural networks} 61: 85--117.

\bibitem[{Schulman et~al.(2015{\natexlab{a}})Schulman, Levine, Abbeel, Jordan
  and Moritz}]{schulman2015trust}
Schulman J, Levine S, Abbeel P, Jordan M and Moritz P (2015{\natexlab{a}})
  Trust region policy optimization.
\newblock In: \emph{International Conference on Machine Learning}. pp.
  1889--1897.

\bibitem[{Schulman et~al.(2015{\natexlab{b}})Schulman, Moritz, Levine, Jordan
  and Abbeel}]{schulman2015high}
Schulman J, Moritz P, Levine S, Jordan M and Abbeel P (2015{\natexlab{b}})
  High-dimensional continuous control using generalized advantage estimation.
\newblock \emph{arXiv preprint arXiv:1506.02438} .

\bibitem[{Schulman et~al.(2017)Schulman, Wolski, Dhariwal, Radford and
  Klimov}]{schulman2017proximal}
Schulman J, Wolski F, Dhariwal P, Radford A and Klimov O (2017) Proximal policy
  optimization algorithms.
\newblock \emph{arXiv preprint arXiv:1707.06347} .

\bibitem[{Shah et~al.(2017)Shah, Dey, Lovett and Kapoor}]{airsim2017fsr}
Shah S, Dey D, Lovett C and Kapoor A (2017) Airsim: High-fidelity visual and
  physical simulation for autonomous vehicles.
\newblock In: \emph{Field and Service Robotics}.
\newblock \urlprefix\url{https://arxiv.org/abs/1705.05065}.

\bibitem[{Silver et~al.(2014)Silver, Lever, Heess, Degris, Wierstra and
  Riedmiller}]{silver2014deterministic}
Silver D, Lever G, Heess N, Degris T, Wierstra D and Riedmiller M (2014)
  Deterministic policy gradient algorithms.
\newblock In: \emph{Proceedings of the 31st International Conference on Machine
  Learning (ICML-14)}. pp. 387--395.

\bibitem[{Stachenfeld et~al.(2017)Stachenfeld, Botvinick and
  Gershman}]{stachenfeld2017hippocampus}
Stachenfeld KL, Botvinick MM and Gershman SJ (2017) The hippocampus as a
  predictive map.
\newblock \emph{Nature neuroscience} 20(11): 1643.

\bibitem[{Stadie et~al.(2017)Stadie, Abbeel and Sutskever}]{stadie2017third}
Stadie BC, Abbeel P and Sutskever I (2017) Third-person imitation learning.
\newblock \emph{arXiv preprint arXiv:1703.01703} .

\bibitem[{Sukhbaatar et~al.(2017)Sukhbaatar, Kostrikov, Szlam and
  Fergus}]{sukhbaatar2017intrinsic}
Sukhbaatar S, Kostrikov I, Szlam A and Fergus R (2017) Intrinsic motivation and
  automatic curricula via asymmetric self-play.
\newblock \emph{arXiv preprint arXiv:1703.05407} .

\bibitem[{Sutton(1991)}]{sutton1991dyna}
Sutton RS (1991) Dyna, an integrated architecture for learning, planning, and
  reacting.
\newblock \emph{ACM SIGART Bulletin} 2(4): 160--163.

\bibitem[{Sutton and Barto(1998)}]{sutton1998reinforcement}
Sutton RS and Barto AG (1998) \emph{Reinforcement learning: An introduction},
  volume~1.
\newblock MIT press Cambridge.

\bibitem[{Szita and L{\"o}rincz(2006)}]{szita2006learning}
Szita I and L{\"o}rincz A (2006) Learning tetris using the noisy cross-entropy
  method.
\newblock \emph{Neural computation} 18(12): 2936--2941.

\bibitem[{Tai et~al.(2016)Tai, Li and Liu}]{tai2016deep}
Tai L, Li S and Liu M (2016) A deep-network solution towards model-less
  obstacle avoidance.
\newblock In: \emph{Intelligent Robots and Systems (IROS), 2016 IEEE/RSJ
  International Conference on}. IEEE, pp. 2759--2764.

\bibitem[{Tai et~al.(2017)Tai, Paolo and Liu}]{tai2017virtual}
Tai L, Paolo G and Liu M (2017) Virtual-to-real deep reinforcement learning:
  Continuous control of mobile robots for mapless navigation.
\newblock In: \emph{2017 IEEE/RSJ International Conference on Intelligent
  Robots and Systems (IROS)}. pp. 31--36.

\bibitem[{Tai et~al.(2018)Tai, Zhang, Liu and Burgard}]{tai2018social}
Tai L, Zhang J, Liu M and Burgard W (2018) Socially-compliant navigation
  through raw depth inputs with generative adversarial imitation learning.
\newblock In: \emph{Robotics and Automation (ICRA), 2018 IEEE International
  Conference on}.

\bibitem[{Tamar et~al.(2016)Tamar, Wu, Thomas, Levine and
  Abbeel}]{tamar2016value}
Tamar A, Wu Y, Thomas G, Levine S and Abbeel P (2016) Value iteration networks.
\newblock In: \emph{Advances in Neural Information Processing Systems}. pp.
  2154--2162.

\bibitem[{Thrun et~al.(2005)Thrun, Burgard and Fox}]{thrun2005probabilistic}
Thrun S, Burgard W and Fox D (2005) \emph{Probabilistic robotics}.

\bibitem[{Tobin et~al.(2017)Tobin, Fong, Ray, Schneider, Zaremba and
  Abbeel}]{tobin2017domain}
Tobin J, Fong R, Ray A, Schneider J, Zaremba W and Abbeel P (2017) Domain
  randomization for transferring deep neural networks from simulation to the
  real world.
\newblock In: \emph{Intelligent Robots and Systems (IROS), 2017 IEEE/RSJ
  International Conference on}. IEEE, pp. 23--30.

\bibitem[{Todorov et~al.(2012)Todorov, Erez and Tassa}]{todorov2012mujoco}
Todorov E, Erez T and Tassa Y (2012) Mujoco: A physics engine for model-based
  control.
\newblock In: \emph{Intelligent Robots and Systems (IROS), 2012 IEEE/RSJ
  International Conference on}. IEEE, pp. 5026--5033.

\bibitem[{Tzeng et~al.(2015)Tzeng, Devin, Hoffman, Finn, Abbeel, Levine, Saenko
  and Darrell}]{tzeng2015towards}
Tzeng E, Devin C, Hoffman J, Finn C, Abbeel P, Levine S, Saenko K and Darrell T
  (2015) Towards adapting deep visuomotor representations from simulated to
  real environments.
\newblock \emph{arXiv preprint arXiv:1511.07111} .

\bibitem[{Tzeng et~al.(2014)Tzeng, Hoffman, Zhang, Saenko and
  Darrell}]{tzeng2014deep}
Tzeng E, Hoffman J, Zhang N, Saenko K and Darrell T (2014) Deep domain
  confusion: Maximizing for domain invariance.
\newblock \emph{arXiv preprint arXiv:1412.3474} .

\bibitem[{Van~Hasselt et~al.(2016)Van~Hasselt, Guez and Silver}]{van2016deep}
Van~Hasselt H, Guez A and Silver D (2016) Deep reinforcement learning with
  double q-learning.
\newblock In: \emph{AAAI}, volume~16. pp. 2094--2100.

\bibitem[{Ve{\v{c}}er{\'\i}k et~al.(2017)Ve{\v{c}}er{\'\i}k, Hester, Scholz,
  Wang, Pietquin, Piot, Heess, Roth{\"o}rl, Lampe and
  Riedmiller}]{vevcerik2017leveraging}
Ve{\v{c}}er{\'\i}k M, Hester T, Scholz J, Wang F, Pietquin O, Piot B, Heess N,
  Roth{\"o}rl T, Lampe T and Riedmiller M (2017) Leveraging demonstrations for
  deep reinforcement learning on robotics problems with sparse rewards.
\newblock \emph{arXiv preprint arXiv:1707.08817} .

\bibitem[{Wang et~al.(2016{\natexlab{a}})Wang, Kurth-Nelson, Tirumala, Soyer,
  Leibo, Munos, Blundell, Kumaran and Botvinick}]{wang2016learning}
Wang JX, Kurth-Nelson Z, Tirumala D, Soyer H, Leibo JZ, Munos R, Blundell C,
  Kumaran D and Botvinick M (2016{\natexlab{a}}) Learning to reinforcement
  learn.
\newblock \emph{arXiv preprint arXiv:1611.05763} .

\bibitem[{Wang et~al.(2016{\natexlab{b}})Wang, de~Freitas and
  Lanctot}]{wang2015dueling}
Wang Z, de~Freitas N and Lanctot M (2016{\natexlab{b}}) Dueling network
  architectures for deep reinforcement learning.
\newblock In: \emph{Proceedings of The 33rd International Conference on Machine
  Learning}.

\bibitem[{Wang et~al.(2017)Wang, Merel, Reed, de~Freitas, Wayne and
  Heess}]{wang2017robust}
Wang Z, Merel JS, Reed SE, de~Freitas N, Wayne G and Heess N (2017) Robust
  imitation of diverse behaviors.
\newblock In: \emph{Advances in Neural Information Processing Systems}. pp.
  5326--5335.

\bibitem[{Weber et~al.(2017)Weber, Racani{\`e}re, Reichert, Buesing, Guez,
  Rezende, Badia, Vinyals, Heess, Li et~al.}]{weber2017imagination}
Weber T, Racani{\`e}re S, Reichert DP, Buesing L, Guez A, Rezende DJ, Badia AP,
  Vinyals O, Heess N, Li Y et~al. (2017) Imagination-augmented agents for deep
  reinforcement learning.
\newblock \emph{arXiv preprint arXiv:1707.06203} .

\bibitem[{Williams(1992)}]{williams1992simple}
Williams RJ (1992) Simple statistical gradient-following algorithms for
  connectionist reinforcement learning.
\newblock In: \emph{Reinforcement Learning}. Springer, pp. 5--32.

\bibitem[{Wu et~al.(2017)Wu, Mansimov, Grosse, Liao and Ba}]{wu2017scalable}
Wu Y, Mansimov E, Grosse RB, Liao S and Ba J (2017) Scalable trust-region
  method for deep reinforcement learning using kronecker-factored
  approximation.
\newblock In: \emph{Advances in neural information processing systems}. pp.
  5285--5294.

\bibitem[{Wu et~al.(2018)Wu, Wu, Gkioxari and Tian}]{wu2018building}
Wu Y, Wu Y, Gkioxari G and Tian Y (2018) Building generalizable agents with a
  realistic and rich 3d environment.
\newblock \emph{arXiv preprint arXiv:1801.02209} .

\bibitem[{Wulfmeier et~al.(2015)Wulfmeier, Ondruska and
  Posner}]{wulfmeier2015maximum}
Wulfmeier M, Ondruska P and Posner I (2015) Maximum entropy deep inverse
  reinforcement learning.
\newblock \emph{arXiv preprint arXiv:1507.04888} .

\bibitem[{Yang et~al.(2018)Yang, Liang and Xing}]{yang2018unsupervised}
Yang L, Liang X and Xing E (2018) Unsupervised real-to-virtual domain
  unification for end-to-end highway driving.
\newblock \emph{arXiv preprint arXiv:1801.03458} .

\bibitem[{You et~al.(2017)You, Pan, Wang and Lu}]{you2017virtual}
You Y, Pan X, Wang Z and Lu C (2017) Virtual to real reinforcement learning for
  autonomous driving.
\newblock \emph{arXiv preprint arXiv:1704.03952} .

\bibitem[{Yu et~al.(2018)Yu, Finn, Xie, Dasari, Zhang, Abbeel and
  Levine}]{yu2018one}
Yu T, Finn C, Xie A, Dasari S, Zhang T, Abbeel P and Levine S (2018) One-shot
  imitation from observing humans via domain-adaptive meta-learning.
\newblock \emph{arXiv preprint arXiv:1802.01557} .

\bibitem[{Zhang et~al.(2017{\natexlab{a}})Zhang, Springenberg, Boedecker and
  Burgard}]{zhang2017deep}
Zhang J, Springenberg JT, Boedecker J and Burgard W (2017{\natexlab{a}}) Deep
  reinforcement learning with successor features for navigation across similar
  environments.
\newblock In: \emph{2017 IEEE/RSJ International Conference on Intelligent
  Robots and Systems (IROS)}. pp. 2371--2378.

\bibitem[{Zhang et~al.(2017{\natexlab{b}})Zhang, Tai, Boedecker, Burgard and
  Liu}]{zhang2017neural}
Zhang J, Tai L, Boedecker J, Burgard W and Liu M (2017{\natexlab{b}}) Neural
  slam.
\newblock \emph{arXiv preprint arXiv:1706.09520} .

\bibitem[{Zhang et~al.(2018)Zhang, Tai, Xiong, Liu, Boedecker and
  Burgard}]{zhang2018vr}
Zhang J, Tai L, Xiong Y, Liu M, Boedecker J and Burgard W (2018) Vr goggles for
  robots: Real-to-sim domain adaptation for visual control.
\newblock \emph{arXiv preprint arXiv:1802.00265} .

\bibitem[{Zhang et~al.(2017{\natexlab{c}})Zhang, McCarthy, Jow, Lee, Goldberg
  and Abbeel}]{zhang2017deepimi}
Zhang T, McCarthy Z, Jow O, Lee D, Goldberg K and Abbeel P (2017{\natexlab{c}})
  Deep imitation learning for complex manipulation tasks from virtual reality
  teleoperation.
\newblock \emph{arXiv preprint arXiv:1710.04615} .

\bibitem[{Zhu et~al.(2017{\natexlab{a}})Zhu, Park, Isola and
  Efros}]{zhu2017unpaired}
Zhu JY, Park T, Isola P and Efros AA (2017{\natexlab{a}}) Unpaired
  image-to-image translation using cycle-consistent adversarial networks.
\newblock \emph{arXiv preprint arXiv:1703.10593} .

\bibitem[{Zhu et~al.(2017{\natexlab{b}})Zhu, Mottaghi, Kolve, Lim, Gupta,
  Fei-Fei and Farhadi}]{zhu2017target}
Zhu Y, Mottaghi R, Kolve E, Lim JJ, Gupta A, Fei-Fei L and Farhadi A
  (2017{\natexlab{b}}) Target-driven visual navigation in indoor scenes using
  deep reinforcement learning.
\newblock In: \emph{Robotics and Automation (ICRA), 2017 IEEE International
  Conference on}. IEEE, pp. 3357--3364.

\bibitem[{Zhu et~al.(2018)Zhu, Wang, Merel, Rusu, Erez, Cabi, Tunyasuvunakool,
  Kram{\'a}r, Hadsell, de~Freitas et~al.}]{zhu2018reinforcement}
Zhu Y, Wang Z, Merel J, Rusu A, Erez T, Cabi S, Tunyasuvunakool S, Kram{\'a}r
  J, Hadsell R, de~Freitas N et~al. (2018) Reinforcement and imitation learning
  for diverse visuomotor skills.
\newblock \emph{arXiv preprint arXiv:1802.09564} .

\bibitem[{Ziebart et~al.(2008)Ziebart, Maas, Bagnell and
  Dey}]{ziebart2008maximum}
Ziebart BD, Maas AL, Bagnell JA and Dey AK (2008) Maximum entropy inverse
  reinforcement learning.
\newblock In: \emph{AAAI}, volume~8. Chicago, IL, USA, pp. 1433--1438.

\end{thebibliography}


%    Biography text here.
%\end{IEEEbiography}
%    Biography text here.
%% insert where needed to balance the two columns on the last page with
%%\newpage
%    Biography text here.


%\vfill


\end{document}